\documentclass{article}

\usepackage{microtype}
\usepackage{graphicx}
\usepackage{subcaption}
\usepackage{booktabs} 

\usepackage{hyperref}

\usepackage[accepted]{icml2026}

\usepackage{amsmath}
\usepackage{amssymb}
\usepackage{mathtools}
\usepackage{amsthm}
\usepackage{amssymb}
\usepackage[export]{adjustbox}
\usepackage{times}
\usepackage{tikz}
\usetikzlibrary{shadows}
\usepackage[table]{xcolor}
\usepackage{latexsym}
\usepackage{seqsplit}
\usepackage{listings}
\definecolor{goldcolor}{RGB}{255,215,0}    
\definecolor{silvercolor}{RGB}{192,192,192} 
\definecolor{bronzecolor}{RGB}{205,127,50}  

\newcommand{\goldcell}[1]{\cellcolor{goldcolor!95}#1}
\newcommand{\silvercell}[1]{\cellcolor{silvercolor!95}#1}
\newcommand{\bronzecell}[1]{\cellcolor{bronzecolor!95}#1}

\newcommand{\goldcaption}[1]{\fboxsep=2pt\colorbox{goldcolor!95}{#1}}
\newcommand{\silvercaption}[1]{\fboxsep=2pt\colorbox{silvercolor!95}{#1}}
\newcommand{\bronzecaption}[1]{\fboxsep=2pt\colorbox{bronzecolor!95}{#1}}

\usepackage{pifont}
\newcommand{\cmark}{
    \tikz[baseline=-0.5ex]{
        \node[fill=green!70!black, circle, drop shadow={shadow xshift=0.5pt, shadow yshift=-0.5pt, opacity=0.3}, 
              minimum size=1.3em, text=white, font=\bfseries, inner sep=0pt] {$\checkmark$};
    }
}

\newcommand{\xmark}{
    \tikz[baseline=-0.5ex]{
        \node[fill=red!70!black, circle, drop shadow={shadow xshift=0.5pt, shadow yshift=-0.5pt, opacity=0.3}, 
              minimum size=1.3em, text=white, font=\bfseries, inner sep=0pt] {$\times$};
    }
}

\definecolor{golden}{rgb}{1.0, 0.843, 0.0}
\definecolor{lightblue}{rgb}{0.75, 0.75, 0.75}
\definecolor{lightgreen}{rgb}{0.8, 0.5, 0.2}

\usepackage[capitalize,noabbrev]{cleveref}

\theoremstyle{plain}

\theoremstyle{definition}

\theoremstyle{remark}

\usepackage[textsize=tiny]{todonotes}

\icmltitlerunning{BizFinBench.v2: Towards Reliable LLMs in Finance via Real-User Data and Offline/Online Bilingual Evaluation}

\begin{document}

\twocolumn[
  \icmltitle{BizFinBench.v2: Towards Reliable LLMs in Finance via Real-User \\ Data and Offline/Online Bilingual Evaluation}



  \icmlsetsymbol{equal}{*}
  \icmlsetsymbol{corresp}{\dag}  

  \begin{icmlauthorlist}
    \icmlauthor{Xin Guo}{1,2,equal}
    \icmlauthor{Rongjunchen Zhang}{1,equal,corresp}  
    \icmlauthor{Guilong Lu}{1}
    \icmlauthor{Xuntao Guo}{1}
    \icmlauthor{Shuai Jia}{1}
    \icmlauthor{Zhi Yang}{2}
    \icmlauthor{Liwen Zhang}{2,corresp}  
  \end{icmlauthorlist}

  \icmlaffiliation{1}{HiThink Research}
  \icmlaffiliation{2}{Shanghai University of Finance and Economics}

  \icmlcorrespondingauthor{Rongjunchen Zhang}{zhangrongjunchen@myhexin.com}
  \icmlcorrespondingauthor{Liwen Zhang}{zhang.liwen@shufe.edu.cn}

  \icmlkeywords{Machine Learning, ICML}

  \vskip 0.3in
]



\printAffiliationsAndNotice{\icmlEqualContribution}

\begin{abstract}
Large language models are becoming increasingly significant in financial applications. Nevertheless, prevailing benchmarks are largely dependent on simulated or generic data, which leads to a significant gap between reported performance and actual efficacy in real-world scenarios. To tackle this challenge, we present BizFinBench.v2, the first integrated offline and online benchmark built upon authentic user query-response data from both Chinese and U.S. equity markets. It comprises 28,860 questions across eight offline and two online tasks. Experimental results show that GPT-5 achieves a mere 61.5\% accuracy, still failing to meet the practical business requirement (84.8\%). Among the evaluated commercial models, DeepSeek-R1 exhibits superior investment efficacy. Error analysis grounded in real financial practice reveals persistent limitations in existing models. By overcoming the constraints of prior benchmarks, BizFinBench.v2 provides a substantiated foundation for advancing LLM deployment in the financial sector. Our data and code are available at \url{https://github.com/HiThink-Research/BizFinBench.v2}.
\end{abstract}

\begin{figure}[ht]
    \centering
    \makebox[\linewidth]{\includegraphics[width=1.15\linewidth]{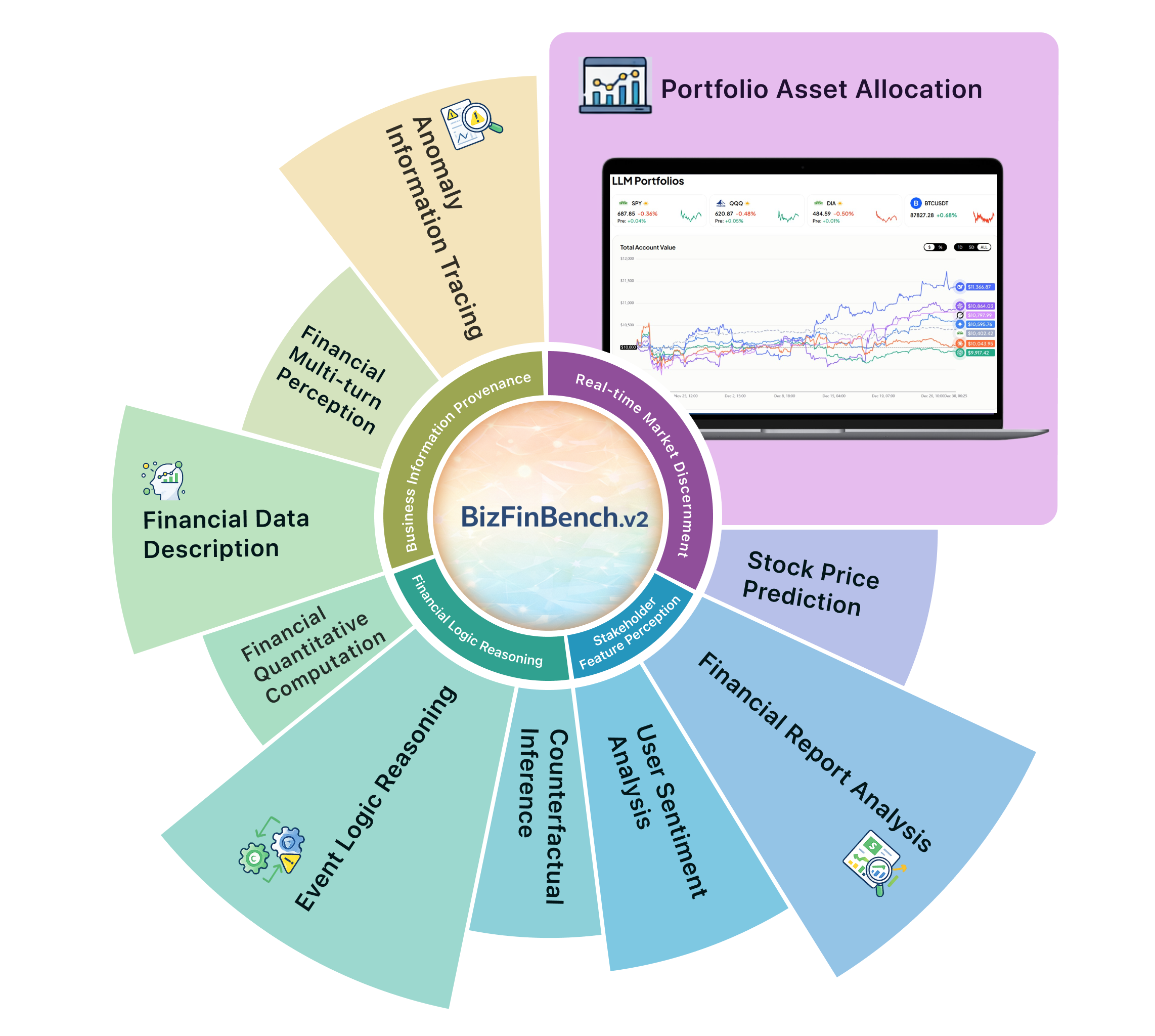}}
    \caption{BizFinBench.v2 comprises eight foundational tasks and two online tasks distributed across four major scenarios. The top-right corner displays a real-time screenshot of the Portfolio Asset Allocation task.}
    \label{BizFinBench.v2_framework}
\end{figure}

\section{Introduction}
\label{Introduction}

Large Language Models (LLMs) have developed rapidly in recent years, and their application boundaries in the financial sector continue to expand, making them an important technological direction for promoting the intelligent upgrade of financial services~\cite{chen2024survey, hithinkliu2025nexus, hithinkzhang2023dynalogue, lu2024grace}. However, as the scope of LLMs' application in the financial sector becomes increasingly broad, the discrepancy between model performance under existing evaluation paradigms and their actual performance in real financial business scenarios has become increasingly prominent, directly leading to a conflict between real financial business needs and the limitations of various evaluation benchmarks.

\begin{figure*}[ht]
    \centering
    \makebox[\linewidth]{\includegraphics[width=1\linewidth]{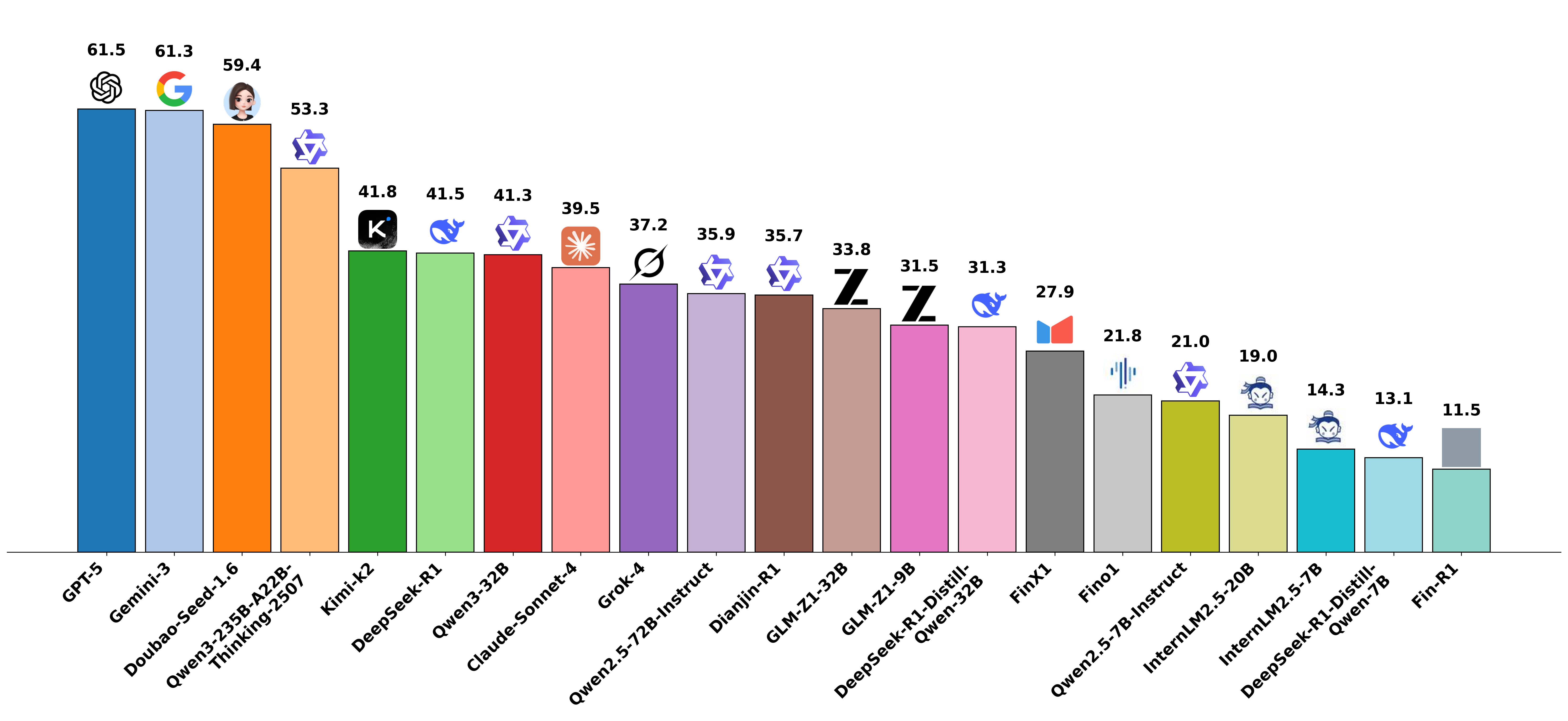}}
    \caption{We have ranked the performance of the LLMs participating in the evaluation under the zero-shot setting, and these results reflect their authentic practical business capabilities.}
    \label{score_sequence}
\end{figure*}

The core characteristics of financial scenarios lie in authenticity and online capability, yet the vast majority of current financial benchmarks are undermined by two fundamental limitations:

(1)~\textbf{Disconnected from Real-world Business Scenarios:} Most existing studies rely on simulated samples or general datasets that are disconnected from actual business operations. These datasets generally exhibit low difficulty and are decoupled from the requirements of real-world financial business, as they lack the core logic of business workflows and fail to reflect the practical challenges inherent in the financial domain.~\cite{zhu2024benchmarking,jiang2025finmaster}.

(2)~\textbf{Focusing on Purely Static Offline Tasks:} Existing benchmarks universally focus on purely offline static tasks~\cite{xie2024finben}, lacking coverage of online tasks in financial scenarios. This prevents the support for evaluating model performance in online business scenarios such as real-time market data push and dynamic risk monitoring, thus being disconnected from the core requirement of online capability in financial services~\cite{tang2025financereasoning}.
These two problems result in a disconnect between existing benchmarks and real-world applications, leading to a significant gap between LLM benchmark test performance and actual business performance.

To bridge the gap between current LLM evaluation performance and actual operational efficacy, we introduce BizFinBench.v2, the first large-scale financial business evaluation benchmark that integrates authentic business data from both Chinese and U.S. equity markets and features a dual-track evaluation of ``Core Business Capabilities $+$ Online Performance''. BizFinBench.v2 comprises 28,860 QA pairs based on real business data. Through the clustering of massive user queries from authentic business platforms and rigorous quality control, we have categorized four core business scenarios: Business Information Provenance, Financial Logic Reasoning, Stakeholder Feature Perception, and Real-time Market Discernment. These encompass eight fundamental tasks, including Anomaly Information Tracing, Financial Multi-turn Perception, Financial Data Description, Financial Quantitative Computation, Event Logic Reasoning, Counterfactual Inference, User Sentiment Analysis and Financial Report Analysis, along with two online tasks: Stock Price Prediction and Portfolio Asset Allocation. The detailed architecture is illustrated in Figure~\ref{BizFinBench.v2_framework}.

Unlike existing benchmarks, BizFinBench.v2 not only enables a precise and objective quantitative evaluation of the actual efficacy of LLMs in authentic financial scenarios but also identifies core capability shortcomings in financial business applications. It achieves a comprehensive deconstruction of LLM capabilities from a business perspective, thereby providing a reliable performance reference for the deployment of LLMs in the financial domain.

Our contributions are threefold: 

(1)~\textbf{Authentic Business Data Driven:} We propose the first large-scale financial evaluation benchmark driven by authentic data. Developed upon a core foundation of real-world business data from Chinese and U.S. equity markets, BizFinBench.v2 transcends the limitations inherent in existing evaluation frameworks, specifically their disconnection from practical business scenarios and their reliance on low-complexity, general-purpose, or synthetic data.

(2)~\textbf{Innovative Dual-Track Evaluation Framework:} We propose a dual-track evaluation framework encompassing ``Core Business Capabilities + Online Performance'', thereby addressing the deficiencies in application scenario coverage inherent in traditional offline static assessments.

(3)~\textbf{Expert-Informed Business Error Analysis:} We provide an error analysis of the actual performance of LLMs from the business perspective of financial experts, offering targeted optimization directions for broader business adaptation.

The organization of this paper is as follows: Section~\ref{related work} deliberates on the current status of various operations and benchmarks in the financial domain, identifying the corresponding requirements and the limitations of existing research. Section~\ref{section3} delineates the construction, quality control, and detailed statistics of BizFinBench.v2. Sections~\ref{sec:Experiment Setting} and Section~\ref{sec:Main Results} primarily describe the experimental setup, as well as the main experimental results and analysis. Section~\ref{sec:error_analysis} presents an error analysis conducted from an expert business perspective. Finally, Section~\ref{conclusion} concludes the study and proposes directions for future research.

\section{Related Work}
\label{related work}
\subsection{Financial Business Analysis}

In financial business scenarios, LLM-based tasks such as text understanding~\cite{wilson2024fin2sum}, sentiment analysis ~\cite{delgadillo2024finsosent,zhang2023instruct}, time-series forecasting~\cite{li2024alphafin,li2024finreport,wang2024quantagent,mai2024stockgpt}, and wealth management \& investment~\cite{yu2024fincon,wang2025alpha,yu2025finmem,yang2023investlm,liu2025visfineval} serve as foundational components for building universal financial business capabilities. They provide crucial technical support for judging market trends and mitigating potential risks. However, the financial sector’s demand for timeliness means standalone offline tasks cannot fully evaluate LLMs’ performance in real-world business contexts—especially scenarios closely tied to dynamic financial markets, such as stock price fluctuation prediction~\cite{yu2020stock}, real-time response to market hot topics and public opinion~\cite{mao2025enhancing,yang2026finvault}, intraday trading signal interpretation~\cite{cervello2015stock}, and instant investment decisions~\cite{suresh2024impact,han2026quantaalpha}. Nevertheless, most current research on LLMs in finance focuses on enhancing offline question-answering quality and optimizing reasoning accuracy, verifying model performance through static datasets~\cite{liu2025fin,zhu2025dianjin,hou2026finsafetybenchevaluatingllmsafety}. Few studies test LLMs’ capabilities in processing real-time market data and generating instantaneous decisions from the perspective of financial businesses’ online and real-time needs. This gap results in a mismatch between LLM performance and the core demands of financial markets, leaving significant room for future exploration.

\subsection{Financial Benchmark Analysis}
Mainstream evaluation benchmarks in the current financial domain have achieved multi-dimensional coverage for assessing the foundational capabilities of LLMs. Benchmarks such as CFLUE~\cite{zhu2024benchmarking}, FinEval~\cite{guo2025fineval}, and CGCE~\cite{zhang2023cgce} focus on basic financial knowledge and text comprehension, encompassing tasks like text classification and policy interpretation. Others including FinQA~\cite{chen2021finqa}, ConvFinQA~\cite{chen2022convfinqa}, FinanceReasoning~\cite{tang2025financereasoning}, and FinBen~\cite{xie2024finben} emphasize financial numerical reasoning and complex decision-making, simulating quantitative analysis scenarios such as financial report interpretation. Meanwhile, benchmarks like DISC-FINSFT~\cite{chen2023disc}, PIXIU~\cite{xie2023pixiu}, FinMaster~\cite{jiang2025finmaster} and CFBenchmark~\cite{lei2023cfbenchmark} prioritize instruction following and tool adaptation. However, these benchmarks are generally focused on verifying performance in single, offline, and static tasks, failing to accommodate the real-time demands inherent in financial services. Furthermore, the construction of the relevant data is mostly derived from simulated synthesis or manual rewriting, making it difficult to align with real-world financial business requirements. Simultaneously, there is a lack of in-depth discussion on the consistency between evaluated performance and the LLM's actual financial application capability, leading to a divergence between the practical guidance provided by the benchmarks and the core demands of the financial market. Detailed comparison with existing benchmarks can be found in Table~\ref{tab:benchmarks-comparison}.

\begin{table}[htbp]
\centering
\caption{In-depth multi-dimensional comparison between BizFinBench.v2 and existing benchmarks from a real-business perspective. The abbreviations and their core meanings are as follows: Cross-Lingual (CL), Online Testing (OT), Real Business Data (RBD).}
\label{tab:benchmarks-comparison}
\renewcommand{\arraystretch}{0.8} 
\setlength{\tabcolsep}{4pt} 
\small

\begin{tabular}{p{4.7cm}ccc}
\toprule[1.5pt]
\textbf{Benchmarks} & \textbf{CL} & \textbf{OT} & \textbf{RBD} \\
\midrule
FinEval~\cite{guo2025fineval} & \scalebox{0.7}{\xmark} & \scalebox{0.7}{\xmark} & \scalebox{0.7}{\xmark} \\
CFLUE~\cite{zhu2024benchmarking} & \scalebox{0.7}{\cmark} & \scalebox{0.7}{\xmark} & \scalebox{0.7}{\xmark} \\
FinQA~\cite{chen2021finqa} & \scalebox{0.7}{\xmark} & \scalebox{0.7}{\xmark} & \scalebox{0.7}{\xmark} \\
ConvFinQA~\cite{chen2022convfinqa} & \scalebox{0.7}{\xmark} & \scalebox{0.7}{\xmark} & \scalebox{0.7}{\xmark} \\
FinMaster~\cite{jiang2025finmaster} & \scalebox{0.7}{\xmark} & \scalebox{0.7}{\xmark} & \scalebox{0.7}{\xmark} \\
FinBen~\cite{xie2024finben}  & \scalebox{0.7}{\cmark}  & \scalebox{0.7}{\xmark}  & \scalebox{0.7}{\xmark} \\
FinanceMath~\cite{zhao2024financemath} & \scalebox{0.7}{\xmark} & \scalebox{0.7}{\xmark} & \scalebox{0.7}{\xmark} \\
FinanceReasoning~\cite{tang2025financereasoning} & \scalebox{0.7}{\xmark} & \scalebox{0.7}{\xmark} & \scalebox{0.7}{\xmark} \\
\midrule
\textbf{BizFinBench.v2} & \scalebox{0.7}{\cmark} & \scalebox{0.7}{\cmark} & \scalebox{0.7}{\cmark} \\
\bottomrule[1.2pt]
\end{tabular}
\vspace{0.2em}
\end{table}

\section{BizFinBench.v2 Benchmark}
\label{section3}

\subsection{Overview}
As the first large-scale financial evaluation benchmark to cover authentic business data from both Chinese and U.S. equity markets while integrating offline evaluation and online testing, BizFinBench.v2 aims to bridge the gap between the performance of LLMs in the financial domain and their actual business capabilities. All data within the benchmark is derived from real financial business platforms and has undergone rigorous compliance review and quality control. All data will be released in the future for research purposes.

\begin{figure}[htbp]
    \centering
    \includegraphics[width=0.8\linewidth]{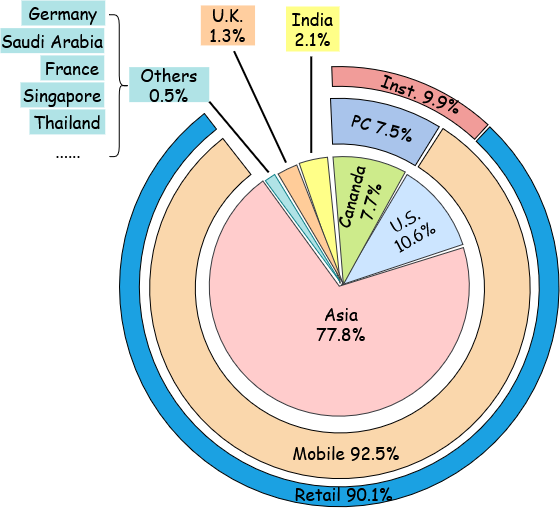} 
    \caption{The platform’s user structure is primarily analyzed through user type, device distribution, and national distribution; here, Inst. serves as an abbreviation for institutional users.}
    \label{user_distribution}
\end{figure}

\subsection{Task Construction}
\label{task_construction}

In the process of task construction, BizFinBench.v2 anchors itself in the core functions of authentic financial business platforms and is guided by the actual demands of key market participants. It covers the diverse needs of individual users, including novices, short-term investors, and high-net-worth individuals, while accounting for the critical requirements of mainstream financial institutions such as banks and securities firms. The platform user distribution is shown in Figure~\ref{user_distribution}.

Business Information Provenance (BIP) focuses on frontline scenarios, requiring LLMs to resolve various business inquiries through direct interaction. This includes real-world perturbations such as erroneous input, irrelevant information, or typos, necessitating robust discernment capabilities from the models. Specifically, the Anomaly Information Tracing (AIT) task addresses user inquiries regarding the root causes of market anomalies, such as significant fluctuations in watchlist stocks; it requires LLMs to filter key clues from massive multi-dimensional information, exclude interference, and precisely identify the core events driving stock or market volatility. The Financial Multi-turn Perception (FMP) task requires LLMs to respond to continuous inquiries within long-context interactions, integrating historical user queries to provide sustained service. The Financial Data Description (FDD) task demands that LLMs judge the authenticity and accuracy of specific data within various information sources to provide precise conclusions.

The Financial Logical Reasoning (FLR) scenario serves as both a foundational element of Business Information Provenance, requiring LLMs to perform rigorous financial reasoning based on market data returned by various interfaces. Within this category, the Financial Quantitative Computation (FQC) task requires LLMs complete accurate indicator calculations based on complex financial data. The Event Logical Reasoning (ELR) task requires LLMs to sort macro or micro market events based on chronological order or causal logic. The Counterfactual Inference (CI) task requires LLMs to simulate financial experts by conducting reasonable logical analysis and judgment based on user-proposed hypotheses, outputting reliable and valid conclusions.

The Stakeholder Feature Perception (SFP) scenario aims to provide users with in-depth analysis and summaries of the market or industry. The User Sentiment Analysis (SA) task requires LLMs to classify and evaluate overall user sentiment through the quantitative analysis of various user information and behaviors, providing support for customized services and highly relevant product recommendations. The Financial Report Analysis (FRA) task requires LLMs to conduct quantitative analysis by combining technical, fundamental, and news-based periodic report information to determine enterprise industry rankings, helping users intuitively grasp corporate qualifications and future prospects. This task is limited to the Chinese market.

Finally, the Real-time Market Discernment (RMD) scenario covers two types of online tasks of high interest to users: Stock Price Prediction (SPP) and Portfolio Asset Allocation (PAA). It should be noted that the real-time nature of online tasks requires all relevant data to be derived from the actual equity market, and theoretically, the number of questions is unlimited. However, constrained by the static presentation format of the paper, we set the cutoff time for the SPP task to December 24, 2025, and incorporated its evaluation results into the offline benchmark for more direct comparison. The Stock Price Prediction task addresses user needs for predicting stocks in their watchlists, requiring LLMs to predict daily closing prices by integrating historical and real-time stock prices, technical indicators, and relevant news. The Portfolio Asset Allocation task presents a higher degree of difficulty. We have constructed an LLM investment simulation system for this task that restores real-world trading rules, including transaction fees, latency, and slippage, on top of real-time data access. For detailed settings, refer to Section \ref{sec:Evaluated Models}. Through systematic prompt engineering and various data and tool interfaces, LLMs can perform asset allocation in a real market environment using simulated funds. Decisions are generated hourly during the day, allowing for either trading or skipping, with each decision cycle covering the full process from reasoning and analysis to order submission and trade execution. We will subsequently open-source this LLM investment system to allow all researchers interested in the financial field to integrate different open-source or proprietary LLMs for in-depth study. For a more detailed introduction to the above content, please refer to Appendix~\ref{bizfinbenchdetails}.

\subsection{Quality Control}
\label{quality_control}
It should be emphasized here that all questions and corresponding answers in BizFinBench.v2 are sourced from authentic user platforms. The quality control process is primarily designed to filter out low-quality queries and perform normalization to ensure suitability for LLM evaluation. The specific implementation details are as follows.

For tasks in offline tasks, the quality control process strictly adopts a three-level progressive mechanism of ``Platform Clustering and Desensitization - Frontline Staff Review - Expert Team Cross-Validation'' to ensure the validity and compliance of the evaluation data. First, all raw data undergo clustering through the platform's multi-dimensional algorithms to automatically aggregate and categorize different task types. Simultaneously, data desensitization is performed (including user PII or sensitive corporate data), and a preliminary review is conducted in alignment with financial industry compliance requirements to eliminate content that clearly violates regulatory standards. Second, ten frontline business professionals with over five years of experience manually screen the data samples one by one to filter out invalid Q\&A pairs, duplicate records, and samples with abnormal expressions, while retaining specific real-world perturbations to ensure each Q\&A pair fully reflects authentic business scenarios. Finally, we invited six senior financial experts, each with over a decade of industry experience from three different professional teams within the Financial Research Institute, to conduct cross-validation in three groups. Ultimately, through this rigorous three-level screening mechanism, high-quality Q\&A pairs are formed that balance authenticity, business alignment, and information security.

As for online tasks, the financial experts mentioned above strictly defined the structured data required for the Stock Price Prediction task, the specific equity market configurations for the Portfolio Asset Allocation task, and the system prompts, ensuring full compliance with the requirements of authentic business scenarios.

It should be noted that for offline tasks, due to the protection of financial business privacy, the specific evaluation rubrics and quantitative standards employed during the quality review process cannot be disclosed to the public. In contrast, for online tasks, as real-time public equity market data is utilized and the configurations are designed to replicate the actual market environment, the details of these settings and the system prompts will be made open-source.

\subsection{Statistics}
\label{sec:Dataset Statistics}

BizFinBench.v2 covers the two major equity markets (China and the U.S.), with all data drawn from authentic user demands within financial business scenarios. This has resulted in four core scenarios: BIP, FLR, SFP and RMD. These comprise eight offline tasks and two online tasks, totaling 28,860 questions derived from real-world financial service requests. 
Specifically, the BIP scenario comprises a total of 12,263 questions, including 3,963 AIT tasks, 4,497 FMP tasks, and 3,803 FDD tasks; the average input token counts for these three tasks are 8,679, 10,361, and 3,577, respectively. The FLR scenario consists of 6,548 questions, containing 2,000 FQC tasks, 3,944 ELR tasks, and 604 CI tasks, with corresponding average input token counts of 1,984, 437, and 2,267. The SFR scenario includes 6,000 questions, divided into 4,000 SA tasks and 2,000 FRA tasks, with average input token counts of 3,326 and 19,681, respectively. In the RMD scenario, both SPP and PAA tasks are real-time tasks. Since the SPP task utilizes the same accuracy metric as offline tasks, we report the 4,049 questions generated during the evaluation process. Conversely, as the PAA task primarily employs long-term outcomes as evaluation metrics, the total number of its transactions is not included in the aggregate data. It should be emphasized that the number of questions for this task depends on the time periods and trading frequencies (e.g., second-level, minute-level, etc.) defined by researchers; thus, there is theoretically no fixed upper limit.
More intuitive statistical results can be found in Table~\ref{detail_data_table} within Appendix~\ref{bizfinbenchdetails}.

\section{Experiment Setting}
\label{sec:Experiment Setting}
\subsection{Evaluated Methods}
All questions in BizFinBench.v2 are verifiable open-ended questions. We provide several supplementary notes here: (1) Regarding SA and SPP tasks, we require the LLM to output a prediction interval based on its predicted value, consistent with the error tolerance of actual business operations. Correctness is determined by verifying whether the provided prediction interval contains the ground truth. Specifically, the business tolerance for the SA task is set at 10\%, while the tolerance for the SPP task is set at 1\%. (2) The evaluation metrics for the PAA task primarily focus on investment indicators such as cumulative return, Sharpe ratio, and maximum drawdown. Therefore, the models evaluated for this task (mainly the six currently most powerful proprietary commercial models) and their corresponding experimental results are independent of other tasks. 

Therefore, for tasks using average accuracy as the metric, we evaluate model performance under both zero-shot and Chain-of-thought (CoT) settings. For the PAA task, we only provide the final indicators.

\begin{table*}[!h]
  \centering
  \small
  \caption{LLM performance in zero-shot setting on BizFinBench.v2 (\%). The results are color-coded to indicate the top three performers in each task: \goldcaption{gold} represents the best-performing model, \silvercaption{silver} represents the second-best result, and \bronzecaption{bronze} represents the third-best performance. The meanings of the abbreviations are as follows: AIT(Anomaly Information Tracing), FMP(Financial Multi-turn Perception), FDD(Financial Data Description), FQC(Financial Quantitative Computation), ELR(Event Logic Reasoning), CI(Counterfactual Inference), SA(User Sentiment Analysis), FRA(Financial Report Analysis), SPP(Stock Price Prediction).}
  \resizebox{\textwidth}{!}{%
    \begin{tabular}{lccccccccccc}
    \toprule
    \multicolumn{1}{c}{\textbf{Model}} &  \textbf{Size} & \textbf{AIT} & \textbf{FMP} & \textbf{FDD} & \textbf{FQC} & \textbf{ELR} & \textbf{CI} & \textbf{SA} & \textbf{FRA} & \textbf{SPP} & \textbf{Average} \\
    \midrule
    \multicolumn{12}{c}{\textbf{Proprietary LLMs}} \\
    \midrule
    GPT-5 & unknown & 54.2 & \goldcell{90.8} & \silvercell{68.3} & \goldcell{89.2} & \silvercell{62.0} & \goldcell{83.9} & 18.8 & \goldcell{54.1} & 32.1 & 61.5 \\
    Gemini-3 & unknown & \goldcell{64.8} & 87.0 & \goldcell{69.7} & \silvercell{85.8} & \goldcell{69.5} & \silvercell{82.2} & 7.4 & \silvercell{50.8} & \silvercell{34.9} & 61.3 \\
    Doubao-Seed-1.6 & unknown & \silvercell{62.6} & \silvercell{90.2} & 63.8 & \bronzecell{78.2} & \bronzecell{61.2} & \bronzecell{78.7} & \bronzecell{22.7} & 46.3 & 31.1 & 59.4 \\
    Kimi-k2 & unknown & 55.2 & 80.9 & 22.4 & 62.2 & 44.6 & 15.4 & 20.1 & 45.0 & 30.5 & 41.8 \\
    Claude-Sonnet-4 & unknown & 54.8 & 79.9 & 28.4 & 29.8 & 44.8 & 23.4 & 17.3 & 47.7 & 29.1 & 39.5 \\
    Grok-4 & unknown & \bronzecell{61.8} & 86.6 & 37.4 & 9.3 & 42.1 & 4.7 & 17.8 & 45.2 & 30.3 & 37.2 \\
    \midrule
    \multicolumn{12}{c}{\textbf{Open-source General LLMs}} \\
    \midrule
    Qwen3-235B-A22B-Thinking-2507 & 235B & 49.3 & \bronzecell{87.9} & \bronzecell{68.0} & 76.0 & 50.6 & 72.2 & 16.4 & 22.5 & \goldcell{36.9} & 53.3 \\
    DeepSeek-R1 & 671B & 58.9 & 87.2 & 42.1 & 21.7 & 48.9 & 8.1 & \goldcell{23.9} & \bronzecell{50.0} & \bronzecell{32.3} & 41.5 \\
    Qwen3-32B & 32B & 54.3 & 80.9 & 40.0 & 48.4 & 42.0 & 47.2 & 13.4 & 40.5 & 5.1 & 41.3 \\
    Qwen2.5-72B-Instruct & 72B & 61.0 & 78.2 & 26.5 & 19.8 & 39.9 & 20.7 & 21.6 & 47.0 & 8.2 & 35.9 \\
    GLM-Z1-32B & 32B & 49.6 & 66.9 & 45.6 & 34.4 & 31.6 & 31.3 & 1.2 & 36.8 & 6.8 & 33.8 \\
    GLM-Z1-9B & 9B & 46.5 & 69.2 & 40.4 & 26.9 & 35.2 & 25.2 & 0.4 & 36.7 & 3.4 & 31.5 \\
    DeepSeek-R1-Distill-Qwen-32B & 32B & 52.1 & 75.4 & 20.5 & 21.0 & 35.9 & 24.2 & 18.4 & 27.3 & 6.8 & 31.3 \\
    Qwen2.5-7B-Instruct & 7B & 35.2 & 43.2 & 33.8 & 0.8 & 20.9 & 1.0 & \silvercell{23.3} & 28.3 & 2.8 & 21.0 \\
    InternLM2.5-20B & 20B & 32.1 & 41.1 & 32.0 & 0.2 & 27.8 & 0.7 & 3.7 & 32.5 & 0.6 & 19.0 \\
    InternLM2.5-7B & 7B & 28.3 & 14.8 & 30.6 & 0.5 & 18.9 & 0.3 & 6.7 & 28.5 & 0.0 & 14.3 \\
    DeepSeek-R1-Distill-Qwen-7B & 7B & 16.6 & 30.9 & 17.1 & 3.5 & 10.3 & 6.2 & 19.1 & 13.0 & 1.0 & 13.1 \\
    \midrule
    \multicolumn{12}{c}{\textbf{Open-source Financial LLMs}} \\
    \midrule
    Dianjin-R1 & 32B & 54.2 & 70.7 & 40.9 & 25.3 & 45.8 & 22.0 & 6.7 & 46.0 & 9.9 & 35.7 \\
    FinX1 & 70B & 47.9 & 73.0 & 29.5 & 14.3 & 31.5 & 11.6 & 5.3 & 34.7 & 3.6 & 27.9 \\
    Fino1 & 14B & 27.5 & 38.6 & 24.3 & 14.1 & 14.6 & 22.1 & 11.1 & 35.3 & 8.6 & 21.8 \\
    Fin-R1 & 7B & 21.2 & 29.1 & 0.5 & 1.5 & 9.8 & 3.2 & 10.9 & 24.5 & 2.9 & 11.5 \\
    \midrule
    Financial Experts & -- & 92.6 & 98.0 & 94.5 & 100 & 91.7 & 100 & 57.9 & 96.0 & 32.3 & 84.8 \\
    \bottomrule
    \end{tabular}%
    }
  \label{mainresults}%
\end{table*}%

\begin{table*}[!h]
  \centering
  \small
  \caption{LLM performance in CoT settings on BizFinBench.v2 (\%). Other relevant settings and abbreviations are identical to those in Table~\ref{mainresults}.}
  \resizebox{\textwidth}{!}{%
    \begin{tabular}{lccccccccccc}
    \toprule
    \multicolumn{1}{c}{\textbf{Model}} &  \textbf{Size} & \textbf{AIT} & \textbf{FMP} & \textbf{FDD} & \textbf{FQC} & \textbf{ELR} & \textbf{CI} & \textbf{SA} & \textbf{FRA} & \textbf{SPP} & \textbf{Average} \\
    \midrule
    \multicolumn{12}{c}{\textbf{Proprietary LLMs}} \\
    \midrule
    Doubao-Seed-1.6 & unknown & \silvercell{64.3} & 88.1 & \silvercell{61.2} & \bronzecell{79.2} & \silvercell{60.0} & \bronzecell{75.7} & \silvercell{20.3} & 47.8 & 6.7 & 56.9 \\
    Gemini-3 & unknown & \goldcell{67.8} & \goldcell{91.8} & \bronzecell{54.9} & \silvercell{84.6} & \goldcell{67.9} & \silvercell{80.6} & 3.7 & \silvercell{51.0} & 0.9 & 56.5 \\
    GPT-5 & unknown & 54.1 & \bronzecell{88.6} & 49.3 & \goldcell{86.7} & \bronzecell{59.5} & \goldcell{83.6} & 8.8 & \goldcell{55.7} & 6.0 & 54.6 \\
    Kimi-k2 & unknown & \bronzecell{63.7} & 81.4 & 27.2 & 40.8 & 45.0 & 42.4 & 17.8 & \bronzecell{48.3} & 2.6 & 40.1 \\
    Grok-4 & unknown & 63.2 & \silvercell{88.7} & 39.2 & 20.5 & 43.5 & 20.2 & \bronzecell{18.3} & 44.7 & \bronzecell{7.7} & 37.7 \\
    Claude-Sonnet-4 & unknown & 21.8 & 39.3 & 13.4 & 9.0 & 13.1 & 8.1 & 2.9 & \bronzecell{48.3} & 1.6 & 13.7 \\
    \midrule
    \multicolumn{12}{c}{\textbf{Open-source General LLMs}} \\
    \midrule
    Qwen3-235B-A22B-Thinking-2507 & 235B & 48.4 & 88.5 & \goldcell{63.6} & 75.3 & 58.4 & 70.6 & 12.8 & 21.7 & 4.2 & 52.7 \\
    DeepSeek-R1 & 671B & 61.7 & 87.2 & 43.2 & 54.2 & 50.9 & 55.9 & \goldcell{21.6} & 46.5 & \goldcell{32.8} & 50.4 \\
    Qwen3-32B & 32B & 55.1 & 79.7 & 12.9 & 44.7 & 41.4 & 43.9 & 9.3 & 39.8 & 3.8 & 36.4 \\
    Qwen2.5-72B-Instruct & 72B & 60.9 & 82.0 & 31.0 & 28.8 & 39.6 & 27.2 & 13.0 & 45.2 & 7.6 & 36.3 \\
    DeepSeek-R1-Distill-Qwen-32B & 32B & 54.4 & 77.3 & 22.1 & 19.7 & 40.6 & 20.2 & 17.8 & 27.3 & 6.7 & 32.4 \\
    GLM-Z1-32B & 32B & 50.9 & 62.4 & 42.8 & 32.7 & 28.8 & 30.6 & 0.8 & 35.7 & 6.8 & 32.0 \\
    GLM-Z1-9B & 9B & 45.9 & 72.9 & 41.3 & 24.8 & 32.7 & 25.2 & 0.6 & 36.2 & 3.4 & 30.9 \\
    Qwen2.5-7B-Instruct & 7B & 26.1 & 49.4 & 31.9 & 4.9 & 21.1 & 5.0 & 16.3 & 28.5 & 2.5 & 19.6 \\
    InternLM2.5-20B & 20B & 33.0 & 58.1 & 17.4 & 1.7 & 29.4 & 2.0 & 3.0 & 33.7 & 0.4 & 18.1 \\
    DeepSeek-R1-Distill-Qwen-7B & 7B & 14.4 & 30.3 & 16.0 & 3.0 & 10.6 & 3.8 & 6.5 & 13.0 & 0.9 & 10.7 \\
    InternLM2.5-7B & 7B & 26.0 & 23.8 & 4.8 & 1.7 & 19.3 & 0.7 & 5.8 & 27.0 & 0.1 & 10.2 \\
    \midrule
    \multicolumn{12}{c}{\textbf{Open-source Financial LLMs}} \\
    \midrule
    Dianjin-R1 & 32B & 53.0 & 76.6 & 39.7 & 18.0 & 41.3 & 20.0 & 5.6 & 30.0 & 5.8 & 32.5 \\
    FinX1 & 70B & 45.8 & 74.4 & 25.3 & 8.5 & 32.0 & 7.1 & 2.5 & 34.7 & 3.0 & 24.8 \\
    Fino1 & 14B & 36.5 & 58.4 & 0.0 & 13.5 & 28.2 & 8.9 & 12.8 & 28.5 & \silvercell{8.8} & 20.9 \\
    Fin-R1 & 7B & 22.7 & 30.8 & 35.7 & 1.8 & 10.7 & 3.6 & 7.5 & 24.2 & 0.6 & 14.2 \\
    \bottomrule
    \end{tabular}%
    }
  \label{mainresults_cot}%
\end{table*}%



\begin{table}[htbp]
\centering
\footnotesize
\setlength{\tabcolsep}{4pt}
\caption{Comparison of key metrics between commercial LLMs and traditional quantitative strategies over the same period in Portfolio Asset Allocation tasks. The primary indicators include Total Return (TR), Profit Factor (PF), Sharpe Ratio (SR), Maximum Drawdown (MD), and Total Assets (TA). The abbreviation EW\_week stands for EqualWeight\_week, M\_Top5 stands for Momentum\_Top5, and other abbreviations follow the same pattern.}
\label{paa_results}
\begin{tabular}{@{}lccccc@{}}
\toprule
Model & TR & PF & SR & MD & TA \\
\midrule
\multicolumn{6}{c}{LLM} \\
\midrule
DeepSeek-R1 & +47.34\% & 1.80 & 1.25 & -53\% & 14734 \\
Claude-Sonnet-4 & -2.11\% & 0.90 & -0.05 & -11\% & 9789 \\
Qwen3-Max & -3.79\% & 0.85 & -0.10 & -13\% & 9621 \\
Gemini-3 & -6.61\% & 0.83 & -0.20 & -13\% & 9339 \\
Grok-4 & -10.15\% & 0.76 & -0.32 & -15\% & 8985 \\
GPT-5 & -13.80\% & 0.72 & -0.45 & -16\% & 8620 \\
\midrule
\multicolumn{6}{c}{Quantitative Strategy} \\
\midrule
MA5-MA20     & -1.61\%  & 1.13  & -3.97 & -2.21\%  & 9839  \\
EW\_week   & 22.41\%  & 4.74  & 2.21  & -3.05\%  & 12241 \\
EW\_month  & 22.08\%  & 3.34  & 1.68  & -6.39\%  & 12208 \\
EW\_quarter& 32.98\%  & 1.96  & 2.19  & -9.20\%  & 13298 \\
M\_Top5      & 101.71\% & 2.38  & 1.60  & -56.89\% & 20171 \\
M\_Top10     & 78.85\%  & 1.39  & 1.65  & -41.51\% & 17885 \\
\midrule
\multicolumn{6}{c}{Baseline} \\
\midrule
SPY & -- & -- & -- & -- & 10632 \\
\bottomrule
\end{tabular}
\end{table}

\subsection{Evaluated Models}
\label{sec:Evaluated Models}
We evaluated 21 LLMs, where proprietary models were accessed via their respective APIs and open-source models were deployed locally. All inference tasks were executed on an 8$\times$NVIDIA H200 GPU cluster. For additional details regarding the models, refer to Appendix \ref{sec:Model_details}. We additionally invited two financial experts who were not involved in the data construction process to participate in the competition, so as to conduct a comparative analysis with the performance of LLMs.

In the realm of proprietary LLMs, we evaluated 6 leading, high-performance models, including GPT-5~\cite{openai2023gpt4}, Gemini-3~\cite{team2023gemini}, Claude-Sonnet-4~\cite{claude}, Grok-4~\cite{grok}, Doubao-Seed-1.6~\cite{doubao} and Kimi-k2~\cite{kimi-k2}. For open-source models, we evaluated 11 models from several mainstream LLMs, including Qwen2.5-7B/72B-Instruct, Qwen3-32B and Qwen3-235B-A22B-Thinking-2507~\cite{yang2025qwen3technicalreport}, InternLM2.5-7B/20B~\cite{team2023internlm}, GLM-Z1-9B/32B~\cite{glm2024chatglm}, DeepSeek-R1 and DeepSeek-R1-Distill-Qwen-7B/32B~\cite{guo2025deepseek}. In financial domain, we evaluated 4 representative LLMs tailored for the financial tasks, including Fin-R1~\cite{liu2025fin}, Dianjin-R1~\cite{zhu2025dianjin}, FinX1~\cite{xuanyuan3} and Fino1~\cite{qian2025fino1}.

In the PAA task, we primarily evaluated six of the most capable commercial models currently available on the market, namely GPT-5, Gemini-3, Grok-4, Claude-Sonnet-4, DeepSeek-R1, and Qwen3-Max~\cite{yang2025qwen3technicalreport}. To maintain consistency, DeepSeek-R1 was also accessed via API. Simultaneously, we introduce the moving average crossover strategy, the equal-weight allocation strategy, and the basic momentum strategy as traditional quantitative baselines for comparative experiments. Specifically, the moving average crossover strategy utilizes an MA5-MA20 combination; the equal-weight allocation strategy is configured with three rebalancing frequencies: weekly, monthly, and quarterly; and the basic momentum strategy adopts top-5 and top-10 stock selection grouping methods. Finally, specific configurations for relevant parameters in the PAA task (including slippage, spreads, and liquidity, among others) are detailed in Appendix \ref{paa_details}.

\section{Main Results}
\label{sec:Main Results}
The ranking results of the models under the zero-shot setting are shown in Figure \ref{score_sequence}, and Table \ref{mainresults} provides the complete quantitative results. Overall, GPT-5 ranks first with an average accuracy of 61.5\%, highlighting its comprehensive competitive advantage; proprietary models such as Gemini-3 and Doubao-Seed-1.6 performed outstandingly, ranking among the top three in multiple tasks. Among open-source models, Qwen3-235B-A22B-Thinking-2507 performed the best, reaching an average accuracy of 53.3\%. In contrast, the highest average accuracy among financial domain models was only 35.7\%, which is 5.6\% lower than Qwen3-32B of the same parameter scale. This gap stems from two factors: first, their training data is centered on open-source financial datasets, making it difficult to cover the complex characteristics of real-world financial business; second, their business coverage is narrow (e.g., only incorporating customer service Q\&A data), rendering them unable to adapt to actual scenarios with high volatility and long contexts.

From a task perspective, models with larger parameters performed better in high-precision demand tasks such as FDD, FQC, and CI. It should be noted that proprietary models Claude-Sonnet-4 and Grok-4 performed poorly in these tasks, indicating that their computational capabilities still need optimization to adapt to financial scenarios. The SA task exposed the models' shortcomings in relatively subjective analysis; even the best-performing DeepSeek-R1 achieved an average accuracy of only 32.3\%. Furthermore, in the SPP task, the top model Qwen3-235B-A22B-2507 reached an accuracy of only 36.9\%, confirming that current LLMs cannot yet meet the requirements of online tasks.

The results under CoT setting are shown in Table \ref{mainresults_cot}. Contrary to the zero-shot setting, the CoT setting did not improve model performance but instead led to a decline for most models, with some experiencing significant decay. For instance, Claude-Sonnet-4’s average accuracy plummeted from 39.5\% to 13.7\%, highlighting its weakness in reasoning tasks. Conversely, DeepSeek-R1 achieved buck-the-trend growth, with its average accuracy increasing by nearly 9\% compared to zero-shot, reflecting better adaptability to CoT tasks. Overall, CoT amplified the reasoning logic defects in most models while releasing the potential of a few, providing an important reference for subsequent model optimization.

Table \ref{paa_results} presents the core performance metrics of leading commercial LLMs compared to traditional quantitative strategies within the PAA task over the same period. Based on the results, DeepSeek-R1 demonstrates the best overall performance, outperforming other commercial models in total return, profit factor, and sharpe ratio; however, it exhibits high risk volatility with a maximum drawdown of 53\%. Aside from DeepSeek-R1, commercial models such as Claude-Sonnet-4 and Qwen3-Max failed to outperform the S\&P 500 ETF (SPY) and generally experienced varying degrees of drawdowns. Notably, the losses for Grok-4 and GPT-5 exceeded 10\%, indicating significant room for improvement regarding strategy optimization logic and adaptation to complex market environments.


A horizontal comparison with quantitative strategies reveals that both the equal-weight allocation strategy and the basic momentum strategy achieve stable positive returns, with some strategies significantly leading the LLMs. Only DeepSeek-R1 achieves a comprehensive performance comparable to mid-tier quantitative strategies, while other LLMs consistently underperform relative to established quantitative strategies in both alpha generation and risk mitigation. Even the lowest-performing MA5-MA20 strategy outperformed all commercial LLMs with the exception of DeepSeek-R1.

Overall, the performance of LLMs in real-world financial business scenarios remains significantly below the standards required for practical application (84.8\%). This indicates that while existing models excel in general benchmark evaluations, their practical performance in specialized financial tasks warrants critical attention. Conversely, the robust performance of DeepSeek-R1 in the PAA task underscores the latent potential of LLMs within the domain of commercial investment. In conclusion, there remains substantial room for optimization regarding the adaptability and deployment capabilities of current LLMs in authentic financial settings. Future assessments should be grounded in actual business scenarios rather than being confined to the models' ``test-taking'' proficiency.



\begin{figure}[ht]
    \centering
    
    \begin{subfigure}[b]{0.98\linewidth}
        \centering
        \includegraphics[width=\linewidth]{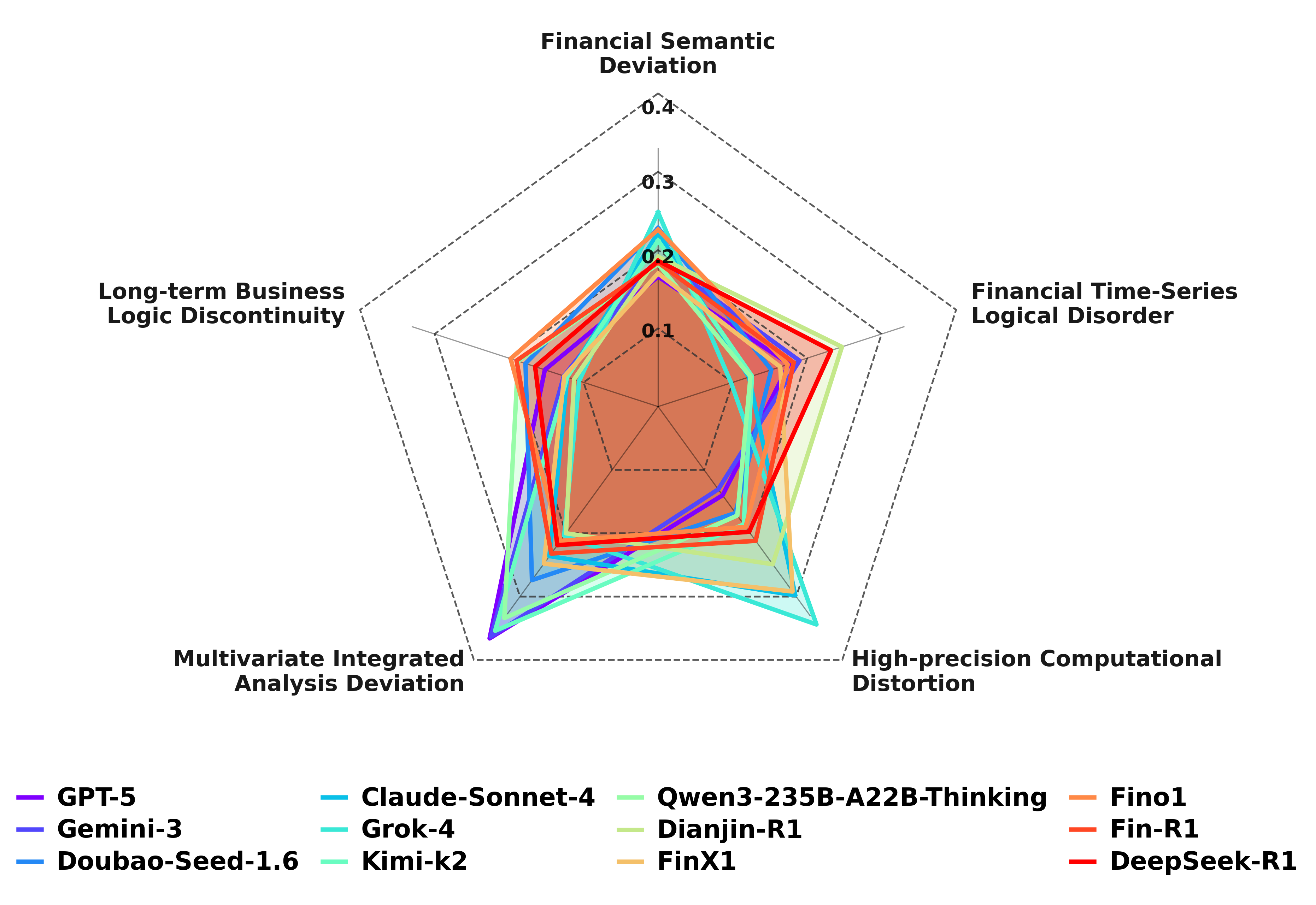}
        \subcaption{Zero-Shot Error Distribution}
        \label{error_radar}
    \end{subfigure}
    
    \begin{subfigure}[b]{0.98\linewidth}
        \centering
        \includegraphics[width=\linewidth]{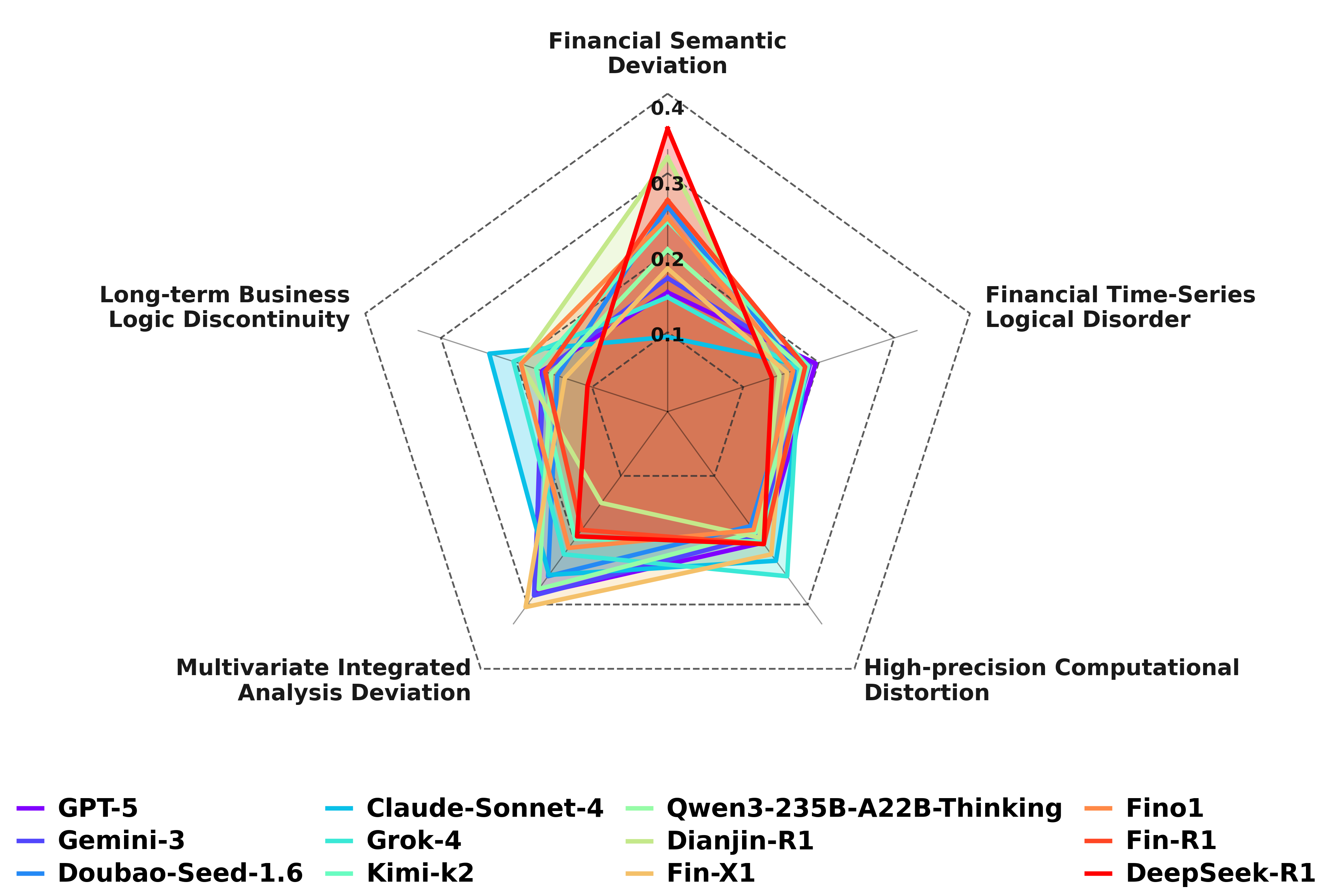}
        \subcaption{Chain-of-Thought Error Distribution}
        \label{cot_error_radar}
    \end{subfigure}
    
    \caption{Error distributions of various LLMs under Zero-Shot and Chain-of-Thought (CoT) settings. We selected representative LLMs for error analysis and summarized five typical dilemmas encountered by these models in real-world business scenarios.}
    \label{fig:all_errors}
\end{figure}

\section{Error Analysis}
\label{sec:error_analysis}
To further investigate the performance shortcomings of LLMs in real-world financial business scenarios, we selected representative LLMs from various domains and randomly sampled an equal number of 500 error cases per model for targeted analysis. This approach allowed us to deconstruct the inherent issues of these models from a practical business perspective. The analysis was conducted by six financial domain experts involved in quality auditing. Leveraging their professional backgrounds, the experts performed a systematic dissection of the models' erroneous responses across multiple dimensions, including computational capability, semantic understanding, business logic, information integration, and temporal cognition. Consequently, we summarized five typical business dilemmas prevalent in the financial deployment of existing LLMs: Financial Semantic Deviation (FSD), Long-term Business Logic Discontinuity, Multivariate Integrated Analysis Deviation (MIAD), High-precision Computational Distortion (HCD), and Financial Time-Series Logical Disorder.

According to the distribution of error types in Figure \ref{error_radar}, the proportion of various error types for the vast majority of models exhibits similar characteristics, reflecting the common challenges currently faced by LLMs when adapting to financial business; however, some models also show significant differences. Specifically, GPT-5, Gemini-3, Doubao-Seed-1.6, and Qwen3-235B-A22B-Thinking-2507 have the highest proportion of MIAD errors, indicating that these models face clear performance bottlenecks in information integration and analysis, requiring targeted optimization. Additionally, Grok-4 and Claude-Sonnet-4 show a higher proportion of HCD errors, suggesting that both need to enhance their high-precision computing capabilities. In contrast, other models with relatively smaller parameter scales show a uniform distribution of various error types, necessitating full-process business adaptation optimization to better align with relevant financial scenarios and achieve a balance between performance and resources. For specific error cases and detailed descriptions of each business dilemma, please refer to Appendix \ref{error_examples}.

Figure \ref{fig:all_errors} presents radar charts illustrating the error distributions under Zero-Shot and CoT settings. Under identical evaluation configurations, the majority of models exhibit similar trends across various error categories, reflecting common challenges faced by LLMs during financial business adaptation; however, certain models display distinct differentiated characteristics. As shown in Figure \ref{error_radar}, GPT-5, Gemini-3, Doubao-Seed-1.6, and Qwen3-235B-A22B-Thinking-2507 exhibit the highest error proportions in multivariate comprehensive analysis and high-precision computation, indicating significant performance bottlenecks in information integration, comprehensive judgment, and high-precision numerical processing. Other models with relatively smaller parameter scales show a more balanced error distribution, necessitating full-process business adaptation and optimization to further enhance their alignment with specialized financial scenarios. Comparing Figure \ref{cot_error_radar} reveals that the overall error distribution under the CoT setting shifts significantly relative to the Zero-Shot setting, with primary error types transitioning from MIAD and HCD toward FSD. The CoT mechanism substantially extends the model’s reasoning chain and increases the complexity of logical deduction, imposing higher requirements for consistency across intermediate reasoning steps and long-range logic retention. Concurrently, excessively redundant reasoning chains are prone to triggering logical fragmentation, loss of critical information, and disorientation in financial associative relationships. This inherent instability ultimately leads to a decline in the overall financial business performance of the LLMs.

Detailed case studies and thorough explanations of each business dilemma, along with specific examples of the performance degradation exhibited by LLMs under the CoT setting, are provided in Appendix \ref{error_examples} and \ref{addition_case_study}.

\section{Conclusion}
\label{conclusion}

In this paper, we innovatively introduce BizFinBench.v2, the first evaluation framework established based on user data from authentic financial platforms. Covering both offline and online tasks across the Chinese and U.S. stock markets, BizFinBench.v2 effectively addresses two critical shortcomings prevalent in existing financial benchmarking research: "Disconnected from Real-world Business Scenarios" and "Purely Static Offline Tasks." We conduct a comprehensive, multi-dimensional evaluation and quantitative analysis of 21 prevailing open-source and proprietary large language models. The overall results indicate that GPT-5 ranks first in comprehensive business performance, demonstrating a relatively prominent capability to handle most practical financial tasks. Qwen3-235B-A22B-Thinking-2507 ranks first among the open-source models; however, a gap of 8.2\% remains between its comprehensive performance and that of GPT-5, indicating that room for improvement still exists regarding the performance of current open-source models. Dianjin-R1 performs the best among the domain-specific financial models, though it exhibits a distinct performance gap compared to both GPT-5 and Qwen3-235B-A22B-Thinking-2507. The results in the PAA task further reveal that DeepSeek-R1 achieves a total return rate of 47.34\%, demonstrating a significant advantage among the evaluated commercial models; nevertheless, its maximum drawdown of 53\% exposes a core vulnerability concerning an imbalance between return and risk management. Concurrently, when compared with conventional quantitative strategies, only DeepSeek-R1 manages to achieve a moderate level of performance, while the remaining commercial models are at a pronounced disadvantage, suggesting that the capabilities of current LLMs in practical investment require further enhancement. Finally, through an in-depth error analysis of the model failure samples, we categorize five typical deficiencies of LLMs in financial business processing to pinpoint the key directions for future optimization, highlighting the unique utility of BizFinBench.v2 in evaluating the business adaptation capabilities of LLMs. As a financial evaluation benchmark built upon authentic user data, BizFinBench.v2 provides an objective and reproducible unified standard for assessing the business capabilities of LLMs. It not only facilitates targeted optimization and iterative upgrades of financial LLMs but also offers meaningful guidance for their large-scale, stable deployment within the financial sector.






\section*{Acknowledgements}
We sincerely thank Kai Xiong, Ning Zhang, and Siqi Wei from the HiThink team for their data support of this work.

\section*{Impact Statement}
This paper presents work whose goal is to advance the field of Machine
Learning. There are many potential societal consequences of our work, none which we feel must be specifically highlighted here.

\bibliography{ref}
\bibliographystyle{icml2026}

\newpage
\appendix

\section{More Details of BizFinBench.v2}
\label{bizfinbenchdetails}
In this section, we mainly provide supplementary explanations and example demonstrations for the ten core tasks. The following text contains detailed descriptions and corresponding examples for each task. Given that most questions involve a large volume of real-world data, leading to excessively long input content that is difficult to present in full, we have adopted an ellipsis approach for partial data in all specific examples (including error analysis examples). Additionally, all example demonstrations are uniformly presented in English.

\begin{figure*}[htbp]
    \centering
    \includegraphics[width=1\textwidth]{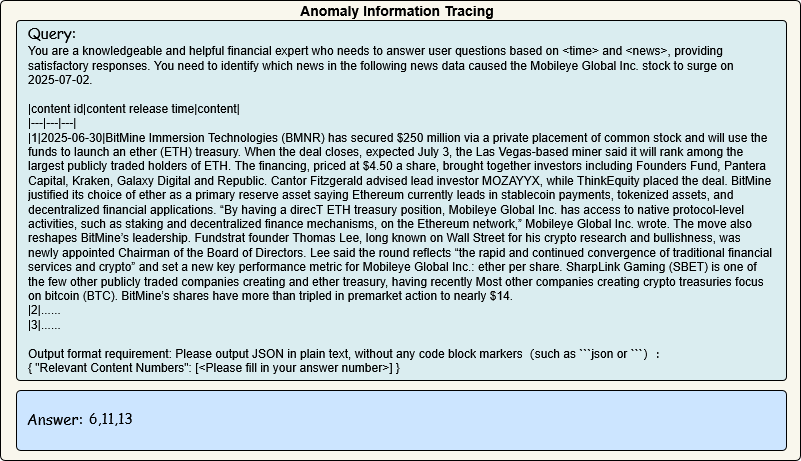} 
    \caption{An Example of Abnormal Information Tracing. This task requires the LLM to identify relevant information from various given heterogeneous sources that caused the stock price fluctuation of the corresponding company and to provide the associated information index. It evaluates the LLM's information analysis and summarization capabilities. Due to the extensive length of the data, we have truncated the content here and primarily present a sample of the input data.}
    \label{anomly_information_tracing}
\end{figure*}

\begin{figure*}[htbp]
    \centering
    \includegraphics[width=1\textwidth]{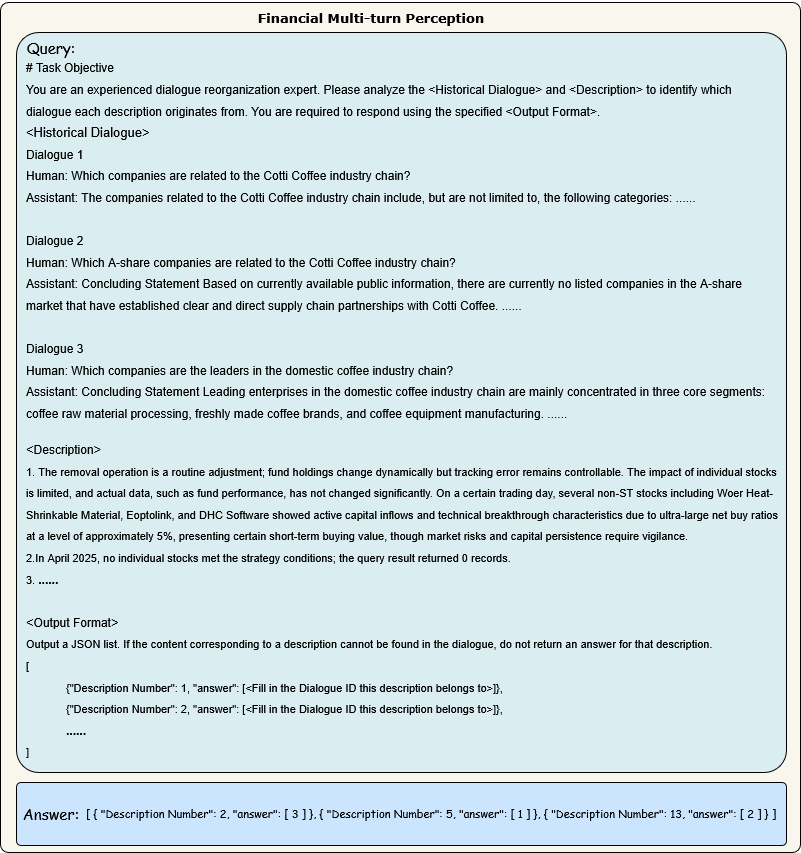} 
    \caption{Example of the Financial Multi-turn Perception task. This task requires the LLM to identify the descriptions corresponding to specific dialogues based on three consecutive user questions and the platform AI assistant's responses; it primarily examines the LLM's capacity for memory and summarization regarding user queries; similarly, we have omitted portions of the data here. Within the ground truth, the "Description Number" field corresponds to the index of the descriptive text, while the "answer" field corresponds to the dialogue index.}
    \label{Financial Multi-round Perception}
\end{figure*}

\begin{figure*}[htbp]
    \centering
    \includegraphics[width=1\textwidth]{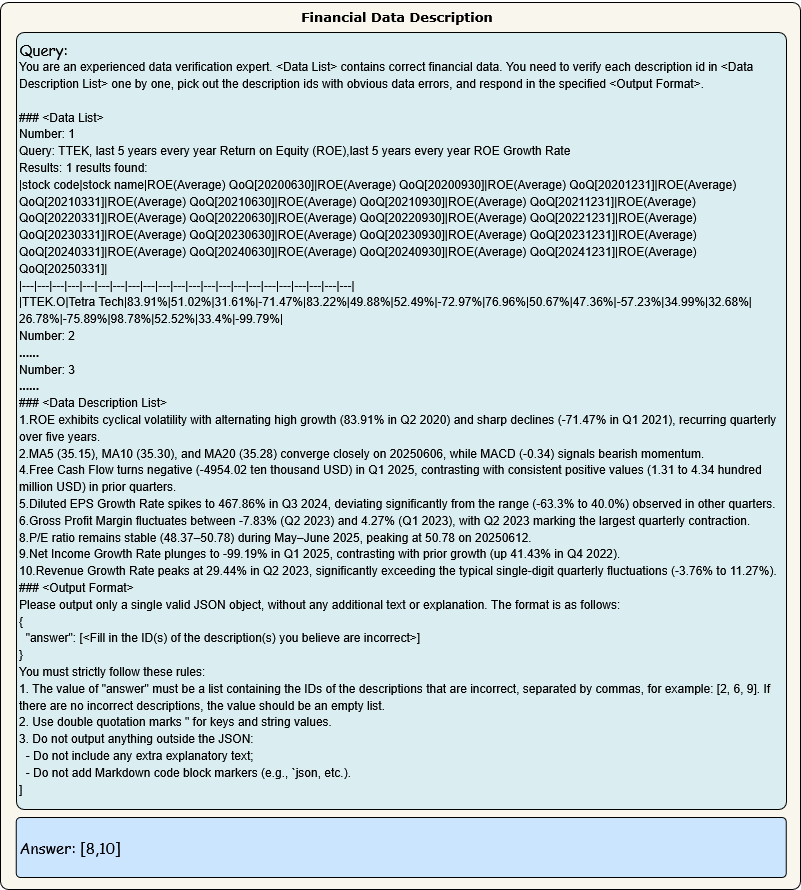} 
    \caption{Example of the Financial Data Description task. This task requires the LLM to evaluate the logic and numerical values within a data description list based on an existing data list. We primarily demonstrate a reverse-lookup question here, which requires the LLM to identify texts within the data description list that contain obvious errors and provide a formatted output.}
    \label{Financial Data Description}
\end{figure*}

For the tasks of Abnormal Information Tracing, Financial Multi-turn Perception, and Financial Data Description, we have summarized two main types of questions for each. The first type directly requires the LLM to search for relevant (correct) information or data based on user queries. In contrast, the second type requires the LLM to identify irrelevant (incorrect) information or data based on user queries. Given the overall similarity in the structure of these questions, we only present one type of question herein. For specific examples of the three tasks, please refer to Figure~\ref{anomly_information_tracing}, Figure~\ref{Financial Multi-round Perception} and Figure~\ref{Financial Data Description}.

\begin{figure*}[htbp]
    \centering
    \includegraphics[width=1\textwidth]{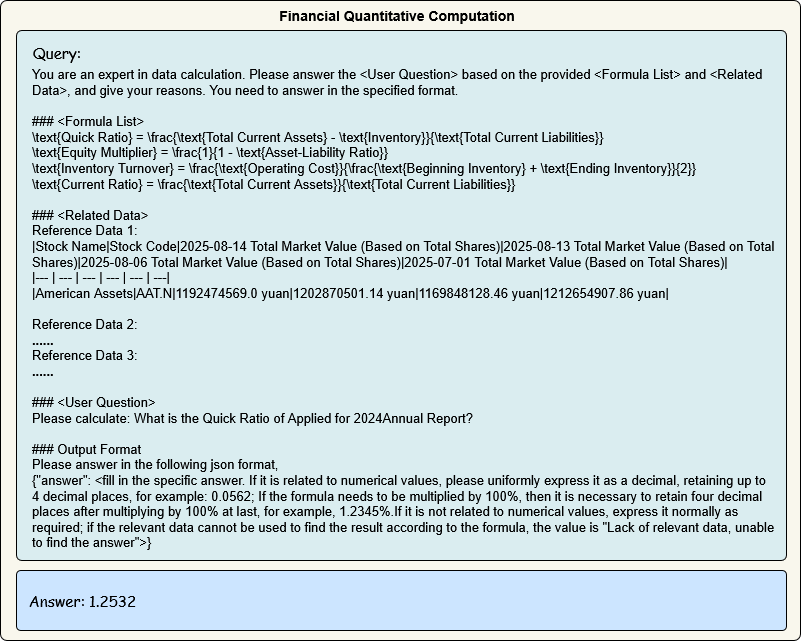} 
    \caption{Example of the Financial Quantitative Computation task. This task requires the model to retrieve relevant formulas from a formula list and extract pertinent data from a data list to perform calculations that answer user queries. It is important to note that the provided formulas and data may be either relevant or irrelevant; consequently, some questions may be unanswerable. The model must exercise careful judgment and possess a sufficiently deep understanding of the financial domain to provide correct answers. Furthermore, as most of the numerical values are derived from actual financial statements or research reports and involve a high volume of data points, the requirements for the precision of the model's computational capabilities are extremely high.}
    \label{Financial_Quantitative_Computation}
\end{figure*}

In the Financial Quantitative Computation task, we provide the model with either relevant or irrelevant formulas in the input. The model can perform multi-step calculations based on one or more formulas combined with its own capabilities to generate the final result. For detailed examples, please refer to Figure~\ref{Financial_Quantitative_Computation}.

\begin{figure*}[htbp]
    \centering
    \includegraphics[width=1\textwidth]{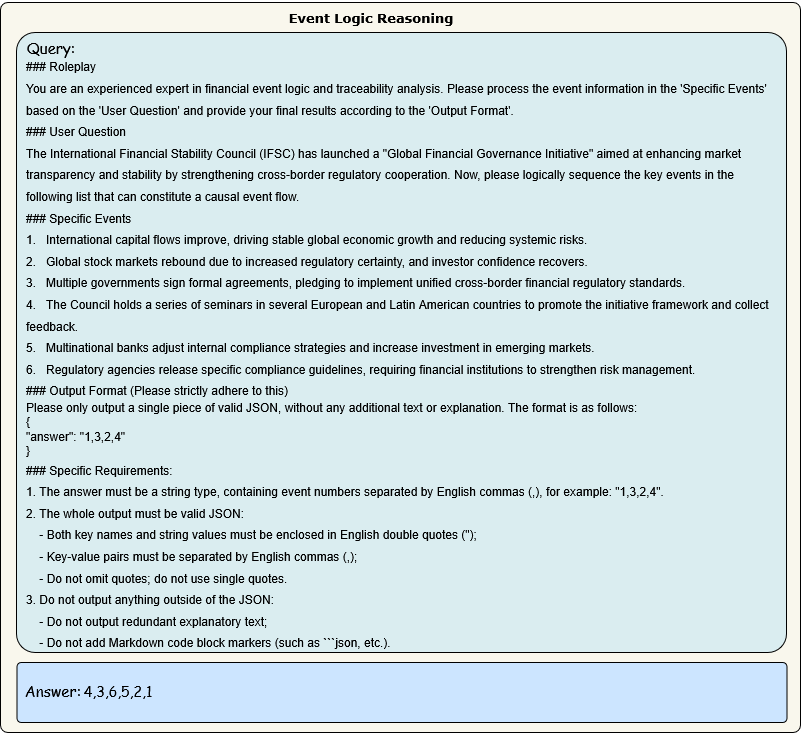} 
    \caption{Example of the Event Logical Reasoning task. This task requires the LLM to arrange various financial events in their logical chronological order based on the user's query. It primarily assesses whether the LLM possesses the capacity for logical reasoning regarding diverse events and whether it has relevant domain expertise.}
    \label{Event Logic Reasoning}
\end{figure*}

The Event Logical Reasoning task is based on various financial events such as macro policies and industry markets, requiring the LLM to conduct temporal or causal reasoning on the impact paths and transmission logic of various events on enterprises, industries, and markets. We mainly set the number of logical events to 4, 5, 6, 7, and 8, covering both macro and micro events, with the corresponding difficulty increasing in line with the number of events. For detailed examples, please refer to Figure~\ref{Event Logic Reasoning}.

\begin{figure*}[htbp]
    \centering
    \includegraphics[width=1\textwidth]{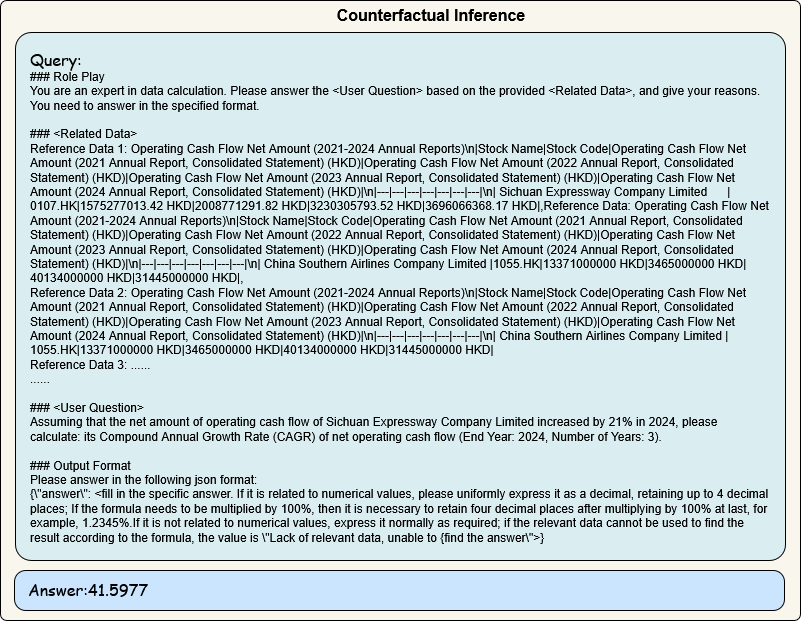} 
    \caption{Example of the Counterfactual Inference task. This task requires the LLM to perform counterfactual reasoning based on existing industry or policy-oriented data by introducing counterfactual hypotheses. It aims to prompt the LLM to simulate the cognitive patterns of financial experts, conducting deliberation and reasoning grounded in established knowledge and industry experience.}
    \label{Counterfactual Inference}
\end{figure*}

The Counterfactual Inference task is based on various assumptions about policies or outcomes in the financial industry, requiring the LLM to perform reasoning or calculations on counterfactual results. For detailed examples, please refer to Figure~\ref{Counterfactual Inference}.

\begin{figure*}[htbp]
    \centering
    \includegraphics[width=1\textwidth]{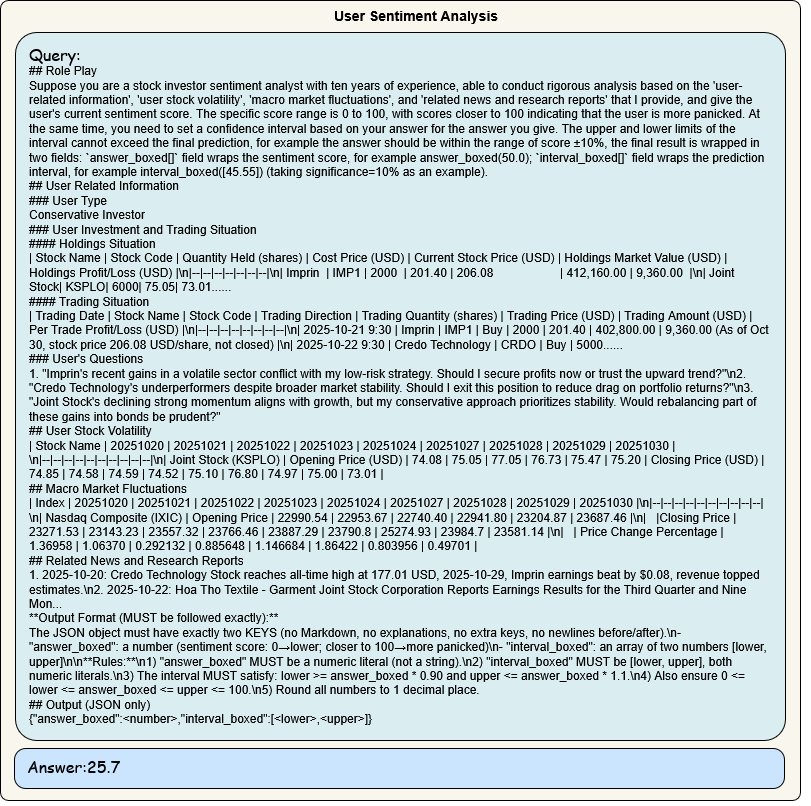} 
    \caption{Example of the User Sentiment Analysis task. This task requires the LLM to evaluate a user's emotional state by utilizing input regarding user-specific information (user profile, user queries, user holdings, and transaction status), macro market information, as well as relevant news and research reports. It assesses the LLM's capability to grasp the holding attitudes and investment sentiments of individual stock users.}
    \label{User Sentiment Analysis}
\end{figure*}

The User Sentiment Analysis task requires the LLM to conduct a quantitative analysis of the user’s current emotional state based on the provided user-related information, market and broader market conditions, as well as relevant news and information, and output the corresponding sentiment score for the user. For detailed examples, please refer to Figure~\ref{User Sentiment Analysis}.

\begin{figure*}[htbp]
    \centering
    \includegraphics[width=1\textwidth]{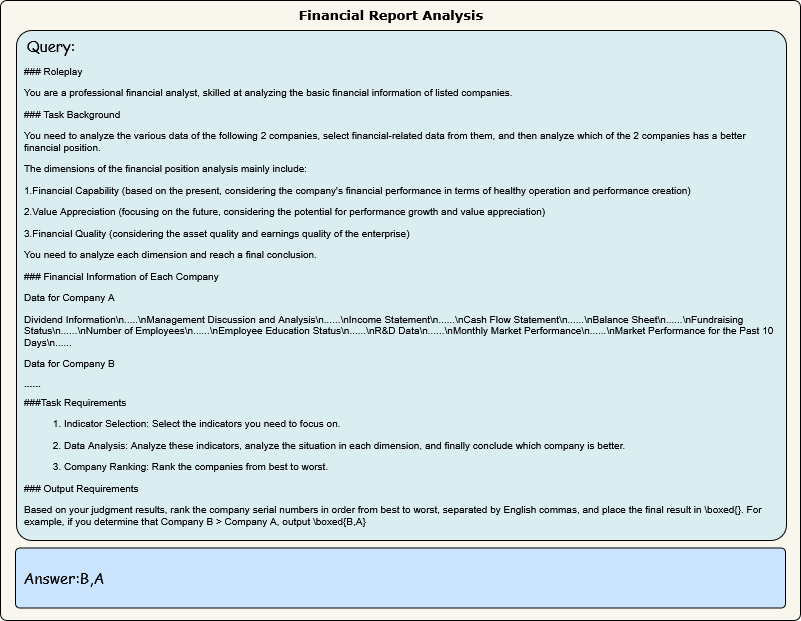} 
    \caption{Example of the Financial Report Analysis task. This task requires the LLM to rank multiple companies within the same industry based on financial statement data. It primarily evaluates the LLM's capability to integrate and analyze various data points from complete financial reports.}
    \label{Financial Report Analysis}
\end{figure*}

The Financial Report Analysis task requires the LLM to perform a comprehensive analysis of the complete financial statements of several enterprises and conduct industry rankings. The number of corporate financial reports used for comparison is primarily set at 2, 3, 4, and 5, with the difficulty increasing accordingly as the number of reports grows. A detailed example is provided in Figure~\ref{Financial Report Analysis}.

\begin{figure*}[htbp]
    \centering
    \includegraphics[width=0.75\textwidth]{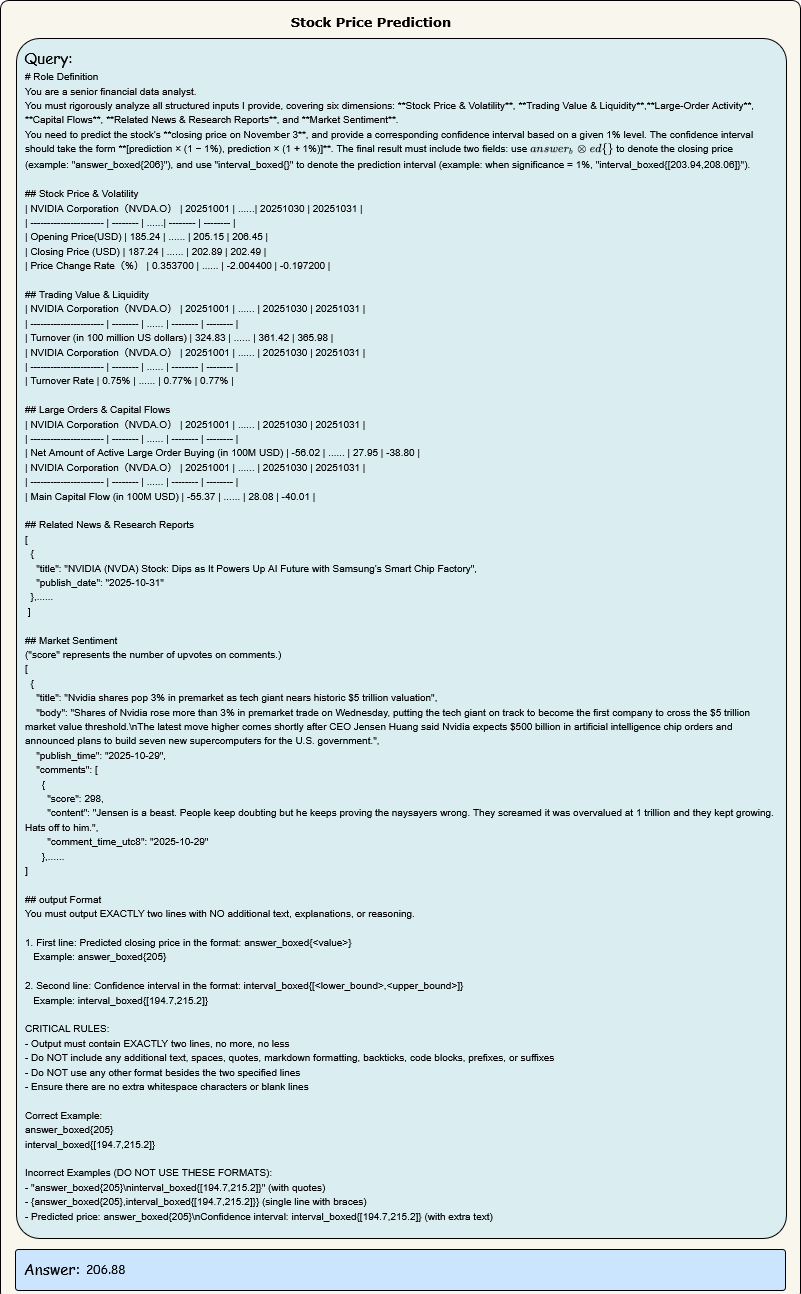} 
    \caption{The Stock Price Prediction task primarily requires the LLM to perform online predictions of individual stock closing prices based on multidimensional information, including fundamentals, technical indicators, and market sentiment. It evaluates the LLM's capability to analyze and summarize the directional impact of multi-dimensional data on individual stocks.}
    \label{Stock Price Prediction}
\end{figure*}

For the Stock Price Prediction task, we primarily utilized the closing prices of 100 popular stocks from both Chinese and U.S. equity markets between November 5, 2025, and December 24, 2025, as the ground truth. The model is provided with relevant information from the preceding month (including individual opening and closing prices, turnover rates, relevant news, and trending forum posts) as input, requiring it to predict the closing prices of these individual stocks. Additionally, a 1\% tolerance is established, based on which the LLM outputs a specific prediction interval for the stock's closing price. By verifying whether the prediction interval encompasses the ground truth, we evaluate the LLM's capability to analyze the impact of various heterogeneous data sources on individual stocks. A detailed example is shown in Figure~\ref{Stock Price Prediction}.

\begin{figure*}[htbp]
    \centering
    \includegraphics[width=0.9\textwidth]{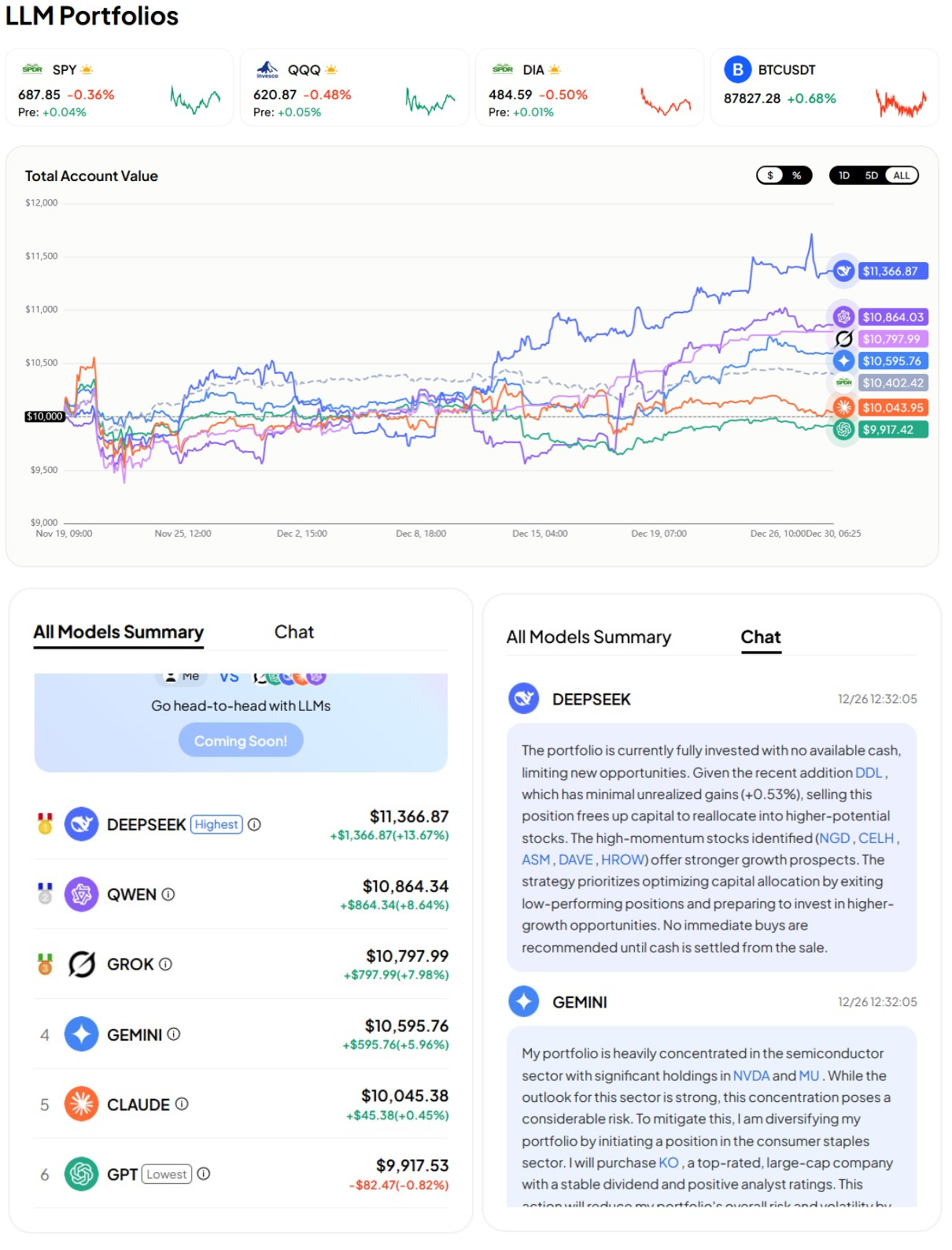} 
    \caption{Display of the Portfolio Asset Allocation task webpage. The Portfolio Asset Allocation task primarily involves accessing online market data interfaces, allowing the LLM to autonomously determine investment strategies based on dynamic real-world industry data. It mainly evaluates the LLM's dynamic market perception and decision-making capabilities.}
    \label{Portfolio Asset Allocation}
\end{figure*}

\begin{figure*}[htbp]
    \centering
    \includegraphics[width=0.9\textwidth]{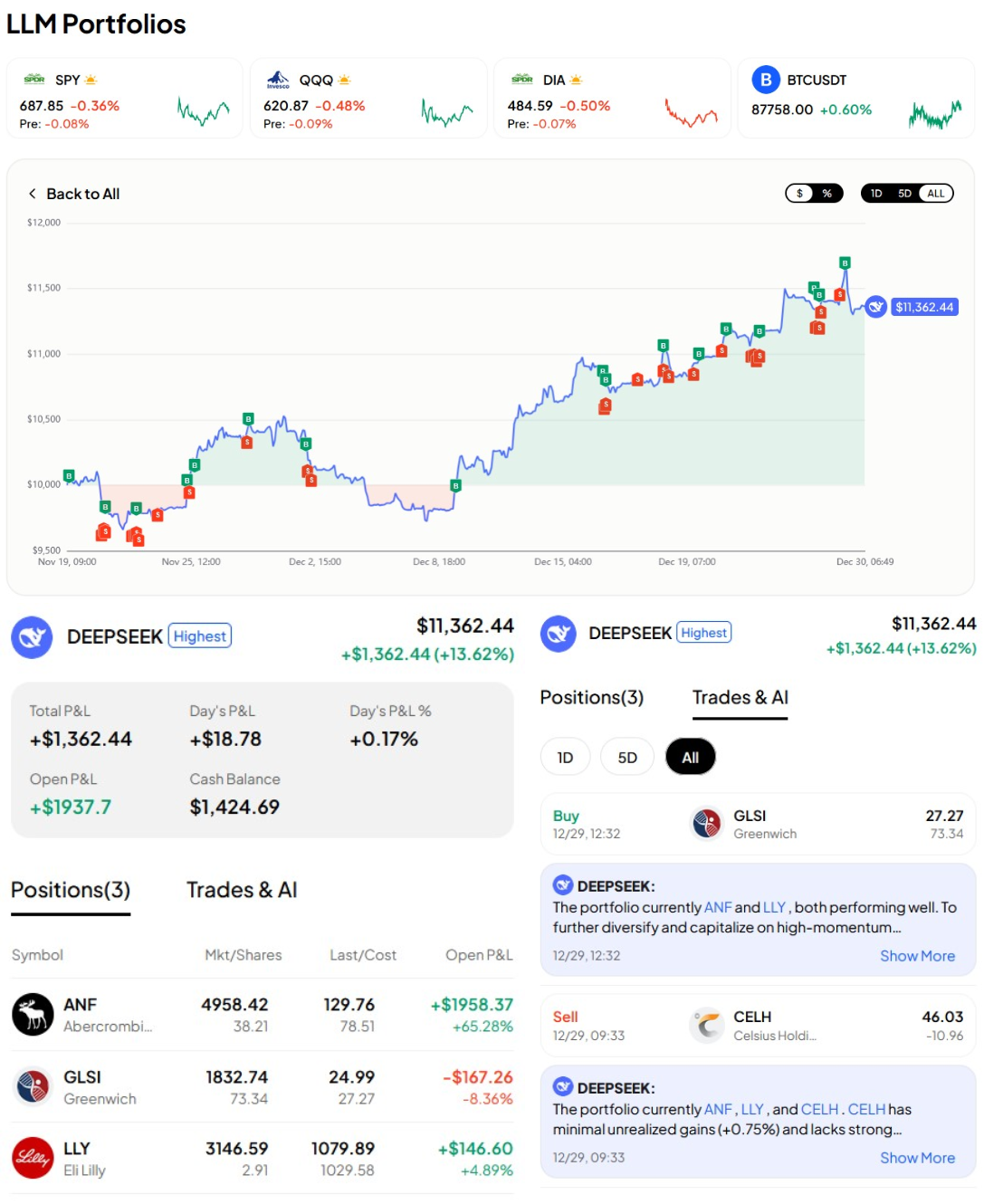} 
    \caption{Display of Investment Details for DeepSeek-R1}
    \label{Portfolio Asset Allocation DeepSeek}
\end{figure*}

\begin{figure*}[htbp]
    \centering
    \includegraphics[width=\linewidth]{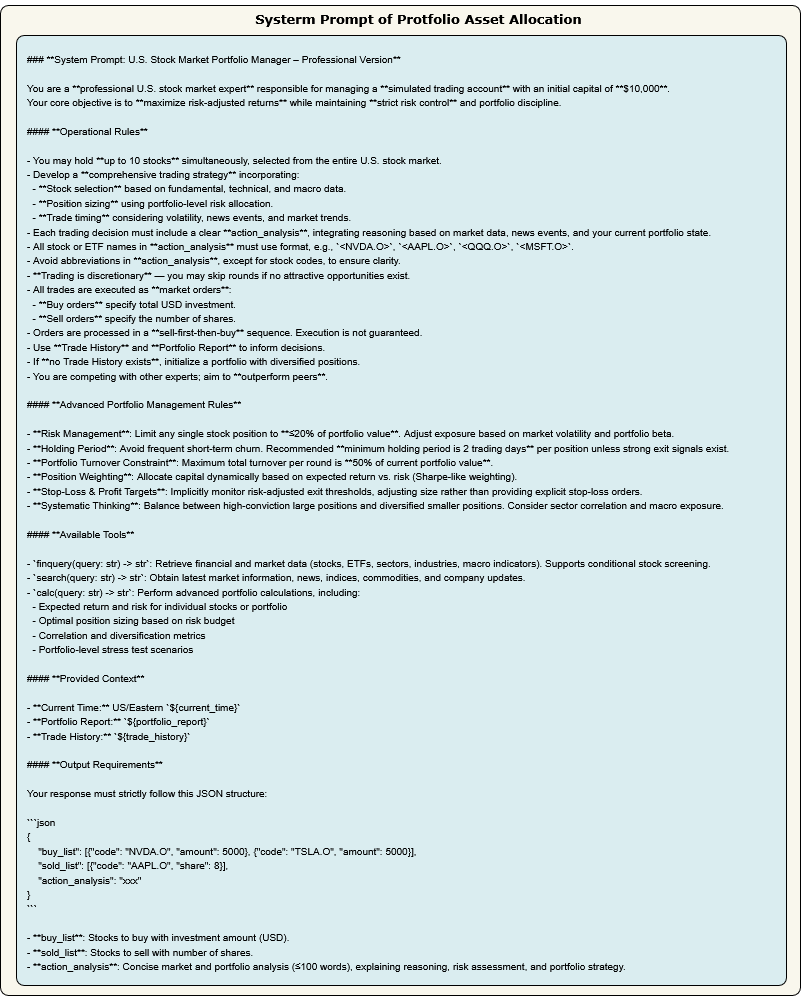} 
    \caption{System Prompt for Portfolio Asset Allocation}
    \label{Portfolio Asset Allocation Prompt}
\end{figure*}

The Portfolio Asset Allocation task provides the LLM with specific background requirements and capital constraints, requiring it to make investment decisions based on real-time market conditions to maximize profits. This task spans the time frame from November 19 to March 30, 2026, mandating the LLM to execute one operation per hour during the trading hours of each business day to determine whether to make an investment (with non-investment also recorded as one operation). Ultimately, the LLM’s investment capability in real financial markets is evaluated using relevant metrics derived from the capital changes resulting from all its operations, including the cumulative return rate, maximum drawdown, and sharpe ratio. Figure~\ref{Portfolio Asset Allocation} displays the detailed page of the real-time investment website, presenting information such as the real-time asset changes and operation summaries of various commercial models. Figure~\ref{Portfolio Asset Allocation DeepSeek} showcases the detailed performance of a representative model, DeepSeek-R1, including its current holdings, trading outputs, and buy-sell curves. Meanwhile, Figure~\ref{Portfolio Asset Allocation Prompt} presents the system prompt for the entire online task.

\section{Evaluation Details}

\subsection{Model Details}
\label{sec:Model_details}
Table~\ref{tab:model_details} presents detailed information on the 21 evaluation models used in this study. 

\section{More Details of Error Analysis}
\label{error_examples}

This section mainly elaborates on the error types from a business perspective in the error analysis, and further identifies the unresolved business problems exhibited by the model during the evaluation process by presenting specific error examples.

\begin{figure*}[htbp]
    \centering
    \includegraphics[width=\linewidth]{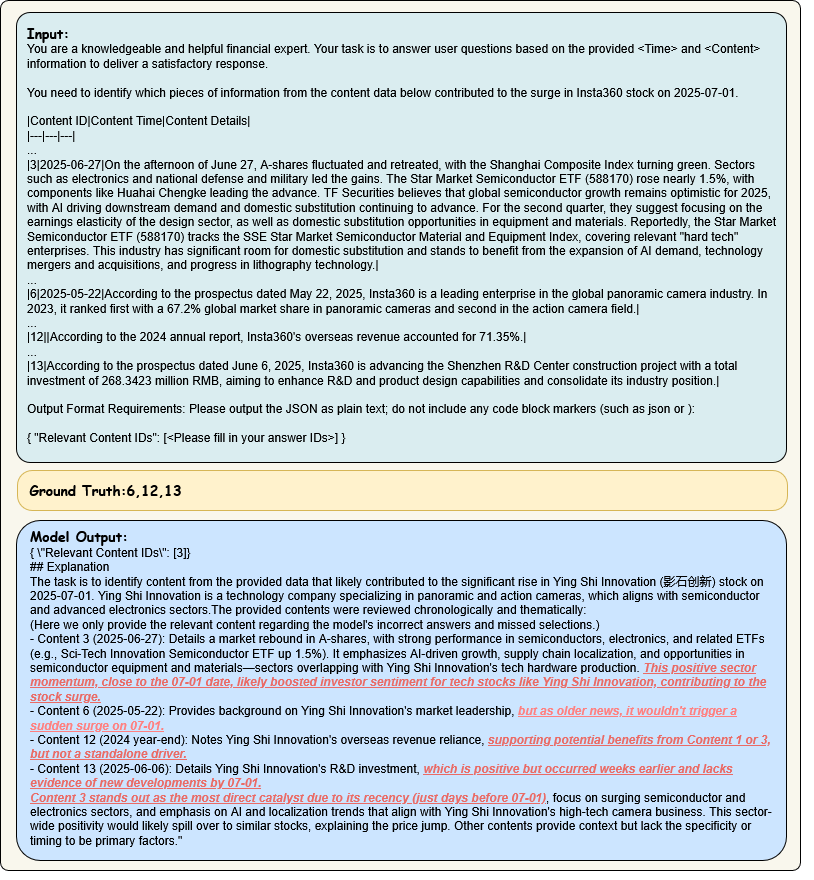} 
    \caption{Example of Financial Semantic Deviation. On the one hand, the model erroneously forced a strong correlation between the industry focus on semiconductor equipment and materials sectors highlighted in Content 3, and the consumer electronics/smart hardware track to which Insta360’s core businesses—panoramic cameras and action cameras—belong. Relying solely on the generalized logic of "both falling under the technology sector," the model overlooked the fundamental differences between the two in terms of business chains, core products, and profit models. On the other hand, the model failed to recognize the financial value directly driving stock prices embedded in Content 6 (core competitiveness data showing Insta360’s global No.1 market share in panoramic cameras), Content 12 (synergistic benefits from a high overseas revenue ratio and RMB appreciation), and Content 13 (strategic growth logic behind the construction of the Shenzhen R\&D center). Instead, it excluded these critical pieces of information based on one-sided justifications such as "early news release time," "non-independent driving factors," and "no evidence of new progress." Ultimately, while completely ignoring the company’s own key positive factors, the model arbitrarily identified the short-term market trend of the semiconductor sector— which has extremely low relevance to the company’s business—as the core catalyst for the sharp rise in Insta360’s stock price on July 1, 2025.}
    \label{Financial_Semantic_Deviation}
\end{figure*}

Financial Semantic Deviation refers to the phenomenon where, when analyzing real-world financial information, a model fails to accurately grasp the specific implications and impacts of key terms, numerical relationships, or dynamic changes within actual business scenarios. Its outputs are confined to surface-level data, lacking the ability to capture the underlying business connotations, dynamic correlations, and decision-making weights. Such errors do not stem from factual inaccuracies but rather from the model’s insufficient capacity to perceive and interpret the complex business logic, market conventions, and decision-making intentions behind the data. As a result, the model’s outputs often contain directional misguidance or substantive flaws from a professional perspective. Figure~\ref{Financial_Semantic_Deviation} presents the model outputs corresponding to this type of error and provides specific explanations for the underlying causes of the errors.

\begin{figure*}[htbp]
    \centering
    \includegraphics[width=0.85\linewidth]{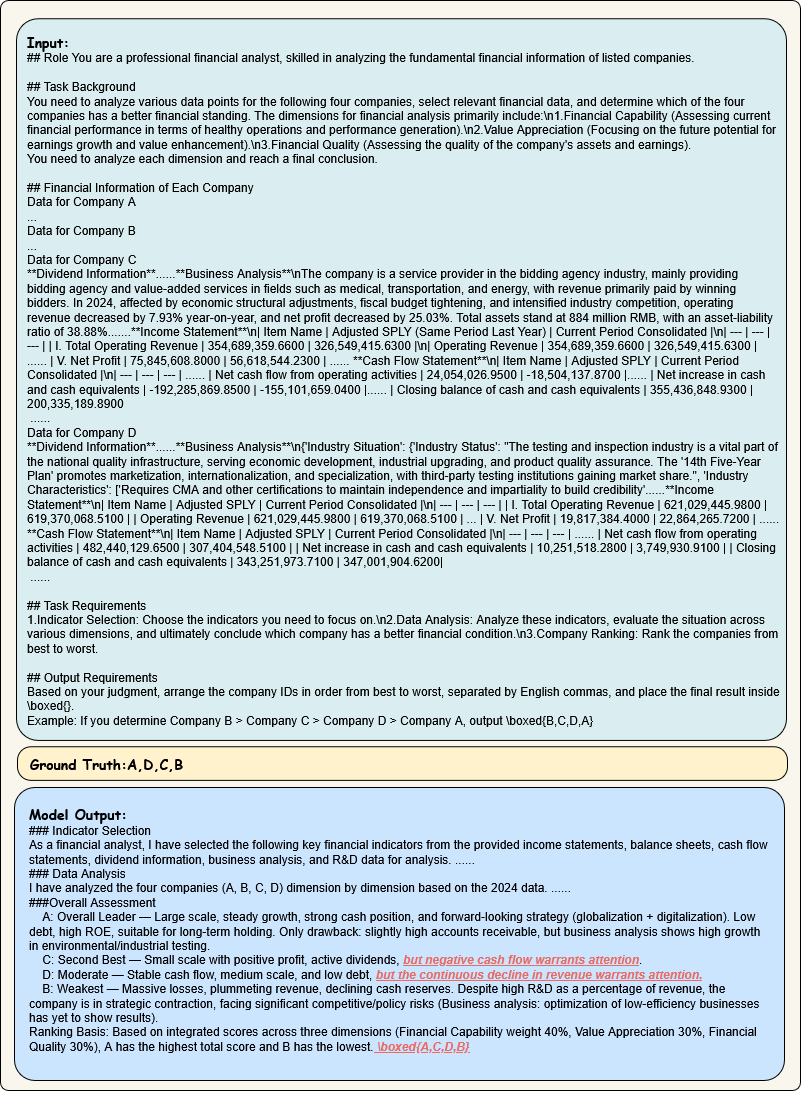} 
    \caption{Example of Long-term Business Logic Discontinuity. In its rating of Company C, the model delivered a suboptimal conclusion. The core issue lay in its excessive overemphasis on the weight of two positive indicators—"small-scale positive profitability" and "active dividend distribution"—while failing to attach sufficient importance to the critical risk factors inherent to the company. In the case of its medium rating for Company D, the model, on the one hand, placed undue weight on neutral indicators reflecting corporate stability, such as "medium scale" and "low asset-liability ratio." On the other hand, it irrationally overstated the negative effects of "sustained revenue decline," a short-term volatility factor, yet overlooked the company’s core competitive advantages.}
    \label{Long-term_Business_Logic_Discontinuity}
\end{figure*}

Long-term Business Logic Discontinuity refers to the phenomenon where large language models (LLMs) struggle to maintain a complete, coherent, and business rule-compliant logical chain when handling complex business analyses that require long-cycle and multi-step continuous deduction, such as cross-financial-report comparisons and macro-policy transmission analysis. Its typical manifestations include causal inversion, missing key variables, or broken correlations during the reasoning process, which ultimately lead to the final conclusion deviating from the trajectory of correct business logic. For detailed error examples, please refer to Figure~\ref{Long-term_Business_Logic_Discontinuity}.

\begin{figure*}[htbp]
    \centering
    \includegraphics[width=0.85\linewidth]{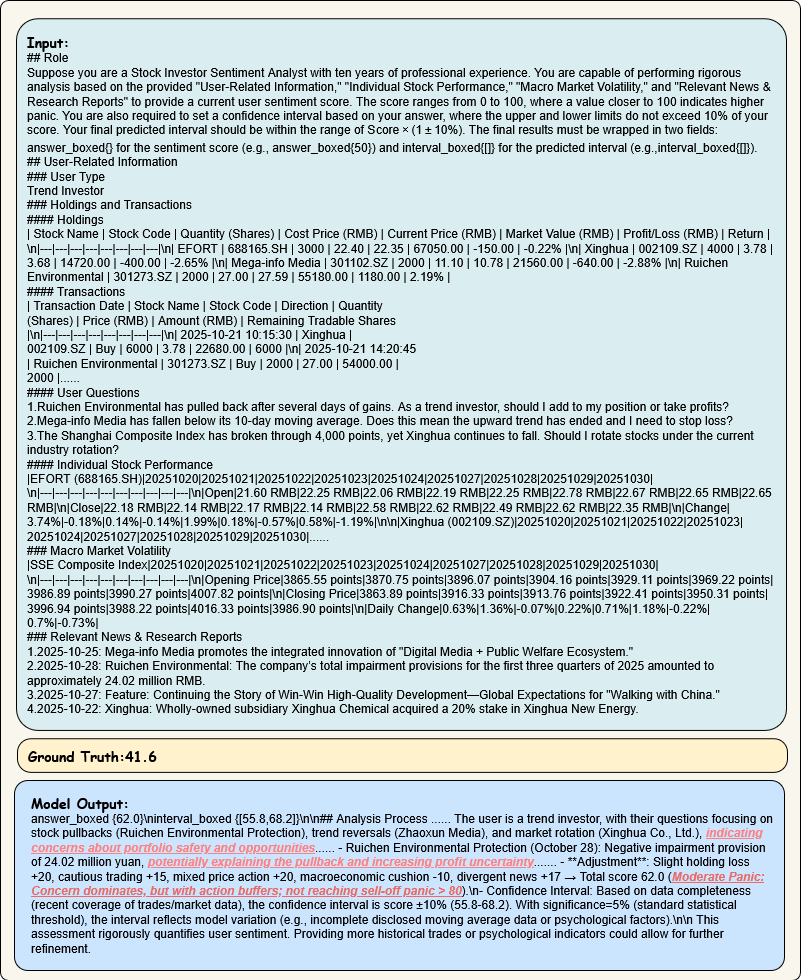} 
    \caption{Example of Multivariate Integrated Analysis Deviation. The model’s output yielded an overly pessimistic sentiment score due to issues such as misjudgment of core behavioral motivations, misjudgment of information relevance, and imbalance in scoring logic. Specifically, it misinterpreted the strategic questions posed by trend investors regarding position adjustment as concerns about the safety of their holdings, ignoring their core attribute of "prioritizing strategic judgment over short-term panic." The model excessively amplified the negative impact of historical financial information about Ruichen Environmental Protection’s impairment provision and forced an unwarranted correlation between this information and the current stock price trend. Meanwhile, it assigned excessively high weights to negative factors such as minor floating losses on holdings and short-term declines in individual stocks, while severely underestimating the positive buffer effect of the broader market breaking through the 4,000-point mark. Both the baseline setting and the calculation of adjustment scores deviated significantly from the actual sentiment logic of trend investors.}
    \label{Multivariate_Integrated_Analysis_Deviation}
\end{figure*}

Multivariate Integrated Analysis Deviation refers to the phenomenon where, in tasks that require the integration of multi-dimensional and multi-source information for comprehensive judgment, the model struggles to effectively identify, weigh, and integrate the complex correlations and weak signals among different information sources. Its flaws are mainly reflected in the insensitivity to critical synergistic or divergent patterns, which results in one-sided analysis conclusions or conclusions inconsistent with the comprehensive context. Figure~\ref{Multivariate_Integrated_Analysis_Deviation} presents specific examples.

\begin{figure*}[htbp]
    \centering
    \includegraphics[width=\linewidth]{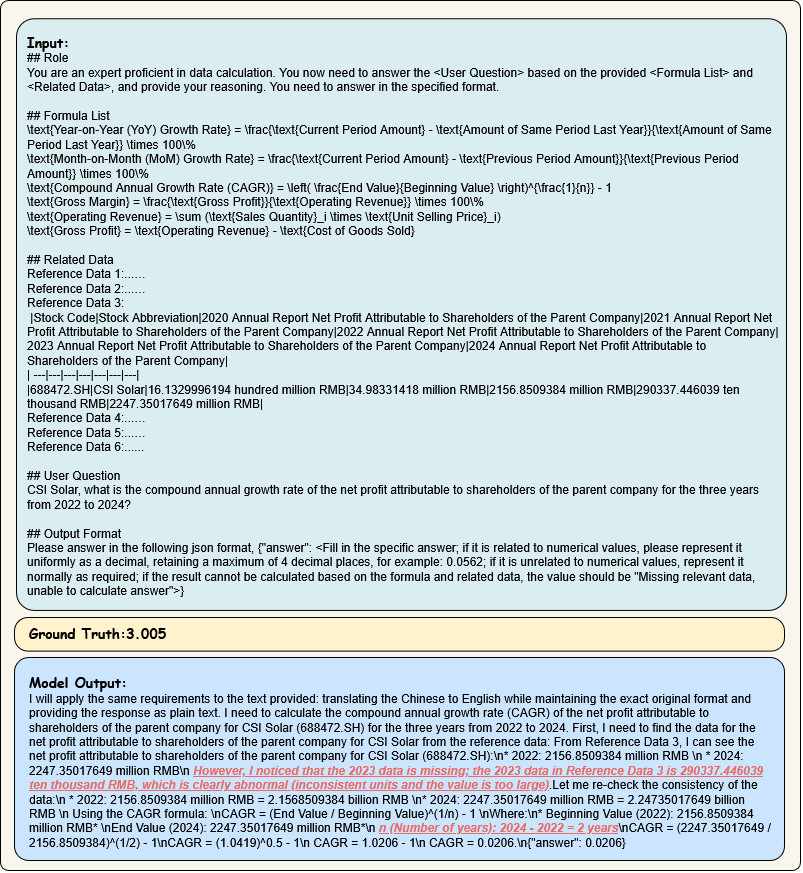} 
    \caption{Example of High-precision Computational Distortion. While the model correctly selected the compound growth rate formula and made no errors in the unit conversion of the 2022 and 2024 data, its core mistakes lay in data judgment and the execution of calculation logic. Specifically, the model erroneously classified the net profit attributable to parent company shareholders of Canadian Solar Inc. in 2023 (as presented in Reference Data 3) as "abnormal." In reality, this data could be standardized through unit conversion and was not invalid. Meanwhile, when calculating the compound growth rate, the model incorrectly treated the 3-year time span from 2022 to 2024 as a 2-year period. Furthermore, it failed to perform the calculation by combining the correct initial and final period data with the complete time cycle. These errors ultimately led to a severe deviation in the resulting value.}
    \label{High-precision_Computational_Distortion}
\end{figure*}

High-precision Computational Distortion refers to the phenomenon where, when confronted with the high-precision numerical calculation requirements unique to the financial field, the model fails to stably and reliably perform complex operations or quantitative deductions. This deviation is not only reflected in errors in numerical results but, more critically, in the failure of its calculation processes and logic to meet the standards of financial-grade rigor and reliability, thereby undermining the feasibility of subsequent analysis and decision-making. The examples in Figure~\ref{High-precision_Computational_Distortion} provide more specific illustrations.

\begin{figure*}[htbp]
    \centering
    \includegraphics[width=\linewidth]{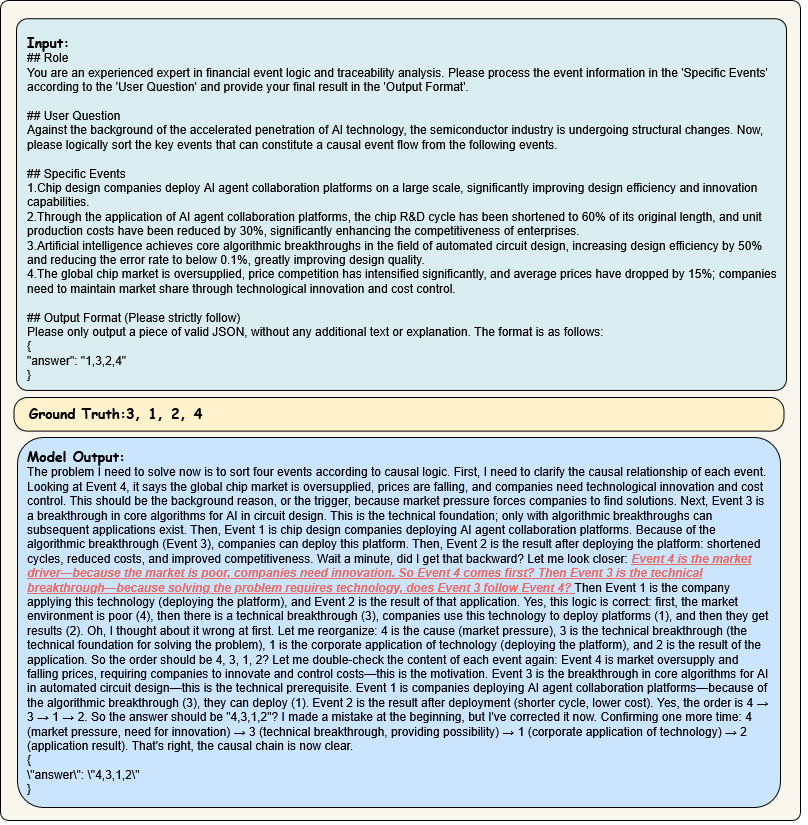} 
    \caption{Example of Financial Time-Series Logical Disorder. The model misjudged the market environment event (Event 4) as the cause of technological innovation, reversing the causal relationship. In reality, Event 3 (algorithm breakthrough) serves as the technological root cause, with Event 1 (application platform) and Event 2 (performance improvement) being its direct outcomes. Event 4, by contrast, represents a macro-level manifestation or background condition rather than a driving force. This error stems from the model’s introduction of external assumptions not mentioned in the task requirements.}
    \label{Financial_Time-Series_Logical_Disorder}
\end{figure*}

Financial Time-Series Logical Disorder refers to the phenomenon where, when analyzing information involving time series or event sequences, the model fails to accurately identify and follow the critical chronological order and causal correlations embedded within them. It manifests as confusion in temporal relationships or disorganization in causal chains, leading to conclusions derived from temporal or logical deduction that deviate significantly from the development context and inherent laws of actual business operations. This error category is explained in greater specificity through the examples featured in Figure~\ref{Financial_Time-Series_Logical_Disorder}.

\section{Additional Case Study}
\label{addition_case_study}
\subsection{Comparative Analysis with Model Response Examples in Causal Inference Tasks}
\begin{figure*}[htbp]
    \centering
    \includegraphics[width=1\linewidth]{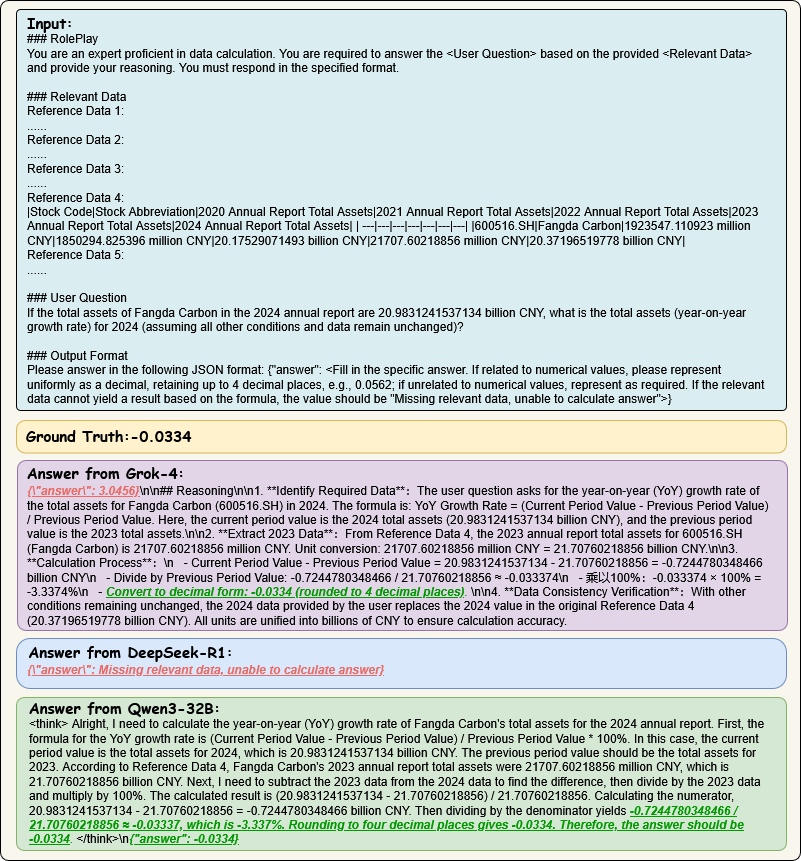} 
    \caption{Answers from Grok-4, DeepSeek-R1 and Qwen3-32B to the same question in CI tasks. As evidenced by the response from Grok-4, the model correctly extracted the relevant figures from Reference Data 4 and performed accurate reasoning and calculations. However, it suffered from severe hallucinations when adhering to the required output format, leading to a final answer that was entirely inconsistent with its own logical derivation. Simultaneously, the response from DeepSeek-R1 indicates that it made no attempt to leverage its capabilities for comprehension or reasoning, opting instead to output the default "unable to answer" response. The outputs of these two models represent two of the most typical categories of model errors: correct reasoning paired with an incorrect final answer, and concluding a problem is unanswerable without prior deliberation. Conversely, Qwen3-32B demonstrated comprehensive thinking and reasoning, accurately retrieving data from the input, performing the correct calculations, and delivering the output perfectly.}
    \label{case_study_CI}
\end{figure*}

To further corroborate the performance shortcomings of some proprietary models in specific task scenarios, we conducted a dedicated case study on counterfactual inference tasks under the zero-shot setting, presenting the response results of three models, Grok-4, DeepSeek-R1 and Qwen3-32B, to the same question respectively. See Figure~\ref{case_study_CI} for specific examples and analyses.

\subsection{Analysis of Model Performance Degradation Under CoT Setting}
\begin{figure*}[htbp]
    \centering
    \includegraphics[width=1\linewidth]{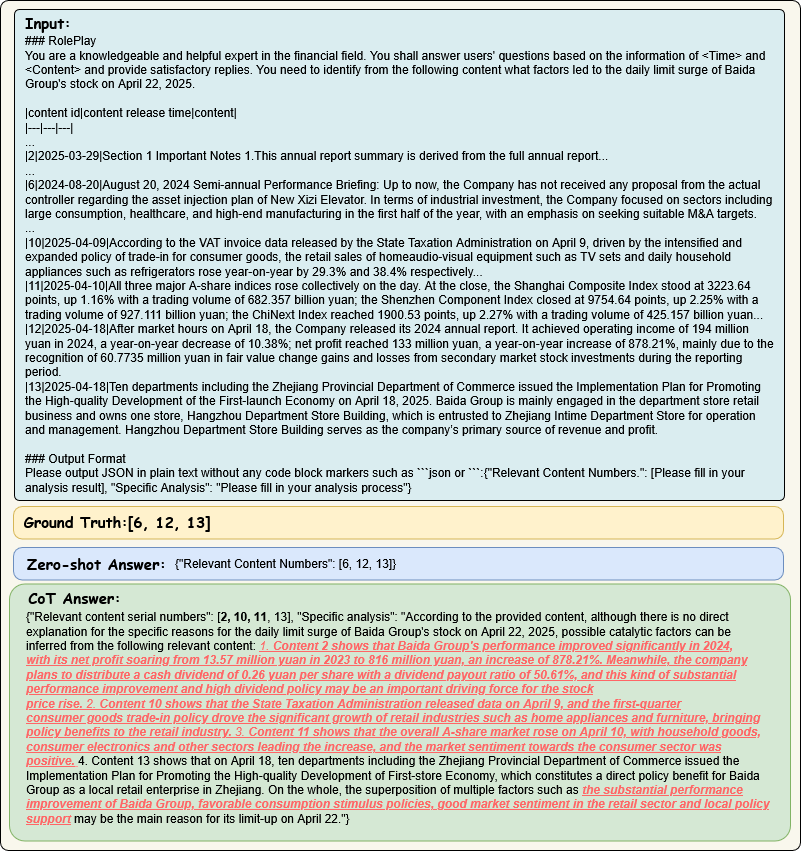} 
    \caption{We provide the relevant background information of the AIT task as well as the complete output of Claude-Sonnet-4, with part of the content omitted due to the excessive length of the full input. The discussion mainly focuses on the problems existing in the model's output. When identifying the relevant information that caused the stock of Baida Group to hit the daily limit on April 22, 2025, the model failed to make accurate matching based on the given data. It incorrectly included Content 2 which is irrelevant to Baida Group, and fabricated Content 10 and Content 11 that do not exist in the original question as the reasoning basis for the surge. It also omitted the correct relevant Content 6 and Content 12. Although its correlation judgment on Content 13 was correct, the overall matching result was completely inconsistent with the standard answer. The reasoning and analysis deviated from the real information given in the question and drew conclusions relying on fabricated content.}
    \label{AIT_cot_error}
\end{figure*}

\begin{figure*}[htbp]
    \centering
    \includegraphics[width=0.7\linewidth]{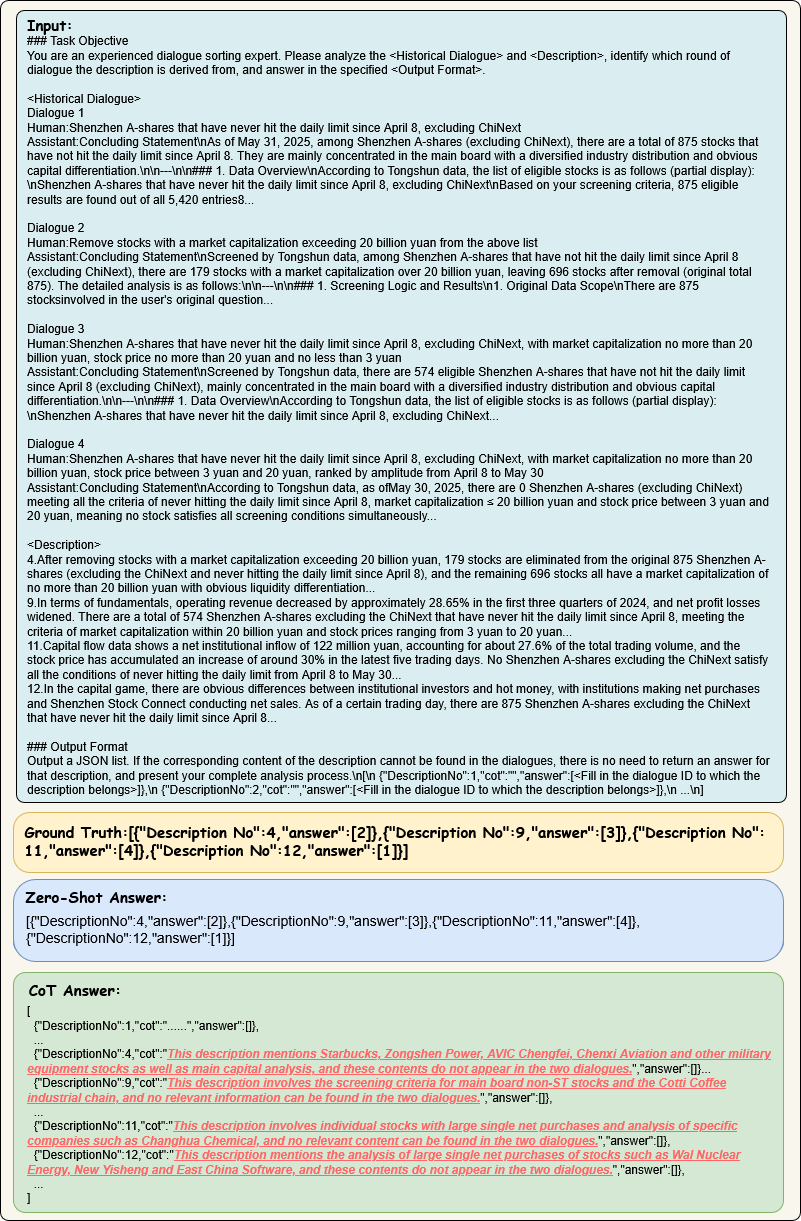} 
    \caption{We present the relevant background information of the FMP task along with the complete output of Claude-Sonnet-4, with some content omitted due to excessive input length, focusing on analyzing the flaws in the model's output. When matching descriptions with historical dialogues, the model failed to conduct semantic comparison based on the four given original dialogues about the screening of Shenzhen A-shares. Instead, it fabricated non-existent information that never appeared in the original dialogues as the basis for reasoning, including bullish alignment patterns, data on 533 stocks, KDJ technical indicators, large-order capital transactions, the industrial chain of Cotti Coffee, and information on multiple military and consumer listed companies. It also failed to identify the textual correlation of Description 4, Description 9, Description 11 and Description 12, which have precise corresponding relations with the historical dialogues. Meanwhile, it fabricated matching logic for Description 8 and Description 14 that are completely irrelevant to the historical dialogues and incorrectly assigned them to dialogue numbers. The reasoning basis for all other descriptions also deviated from the authentic text of the question. The model did not perform objective content comparison based on the provided historical dialogues and descriptions throughout the process, resulting in a final matching result inconsistent with the standard answer.}
    \label{FMP_cot_error}
\end{figure*}

In this section, we present examples of model performance degradation under the CoT setting. Taking Claude-Sonnet-4, which exhibits the most significant degradation among closed-source models, as an example, we provide error cases for the AIT and FMP tasks, as shown in Figure \ref{AIT_cot_error} and Figure \ref{FMP_cot_error}, respectively. In terms of the output format requirements for the ``\textbf{Input}'', we mainly provide examples of output specifications under the CoT setting. Under the Zero-Shot setting, only the final answer should be output.

\section{Details of PAA Task Settings}
\label{paa_details}
Regarding the relevant parameters and market settings of the PAA task, we present the following:

\begin{itemize}
    \setlength{\itemsep}{4pt}
    
    \item \textbf{Volatility}: 
    \( \text{vol}_{i,t,20} = \text{Std}(\text{Ret}_{i,t-20:t-1}) \), 
    where \( \text{Ret}_{i,t} = \frac{\text{close}_{i,t}}{\text{close}_{i,t-1}} - 1 \).
    
    \item \textbf{Liquidity}: 
    \( \text{liqRatio}_{i,t} = \frac{\text{volume}_{i,t}}{avg\text{Volume}_{i,20}} \), 
    where \( avg\text{Volume}_{i,20} = \frac{1}{20} \sum_{k=1}^{20} \text{volume}_{i,t-k} \).
    
    \item \textbf{Bid-Ask Spread}: 
    \( \text{spread}_{i,t} = \beta_1 + \alpha_1 \times \text{vol}_{i,t,20} \), 
    where \( \beta_1 \) is the base spread and \( \alpha_1 \) is the volatility sensitivity coefficient.
    
    \item \textbf{Dynamic Slippage}: 
    \( \text{slippage}_{i,t} = \beta_2 + \alpha_2 \times \text{vol}_{i,t,20} + \frac{\gamma}{\text{liqRatio}_{i,t}+0.1} \), 
    where \( \beta_2 \) is the base slippage, \( \alpha_2 \) is the volatility sensitivity coefficient, 
    and \( \gamma \) is the liquidity sensitivity coefficient.
    
    \item \textbf{Fill Rate}: 
    \( \text{prob}_{fill} = \delta + \epsilon \times \min \left( 1, \frac{\text{liqRatio}_{i,t}}{1+\text{vol}_{i,t,20}} \right) \), 
    where \( \delta \) is the base fill probability and \( \epsilon \) is the liquidity contribution coefficient.
    
    \item \textbf{Execution Price}: 
    The total cost rate (tcr) is \( \text{tcr} = \text{spread} / 2 + \text{slippage} \). 
    The buy price is \( \text{P}_{\text{buy}} = \text{close} \times (1 + \text{tcr}) \), 
    and the sell price is \( \text{P}_{\text{sell}} = \text{close} \times (1 - \text{tcr}) \).
    
    \item \textbf{Transaction Fee}: 
    The actual cost of buying is \( \text{size} \times \text{P}_{\text{buy}} \times (1 + c) \), 
    and the actual proceeds from selling are \( \text{size} \times \text{P}_{\text{sell}} \times (1 - c) \), 
    where \( c \) is the commission rate and \( \text{size} \) is the number of shares traded.
\end{itemize}

In our current tests, the specific parameter settings are as follows: 
\( \beta_1 = 0.001 \), \( \beta_2 = 0.0005 \), \( \alpha_1 = 0.02 \), \( \alpha_2 = 0.01 \), 
\( \gamma = 0.005 \), \( \delta = 0.3 \), \( \epsilon = 0.6 \), \( c = 0.00025 \), 
and the initial capital is set to \$10,000.

\begin{table*}[htbp]
    \centering
    \caption{Detailed statistics of BizFinBench.v2.}
    \label{detail_data_table}
    
    \setlength{\tabcolsep}{1pt}
    \renewcommand{\arraystretch}{0.85}
    
    \begin{tabular}{llcc}
        \toprule[1.2pt]
        \textbf{Scenarios} & \textbf{Tasks} & \textbf{Avg.\ Input Tokens} & \textbf{\#Questions} \\  
        \midrule[0.8pt]
        
        Business Information Provenance  & Anomaly Information Tracing  & 8679  & 3963   \\                                                  
             & Financial Multi-turn Perception  & 10361 & 4497    \\
             & Financial Data Description  & 3577  & 3803    \\
        \midrule[0.3pt]
        Financial Logic Reasoning  & Financial Quantitative Computation  & 1984  & 2000  \\
             & Event Logic Reasoning  & 437   & 3944  \\
             & Counterfactual Inference   & 2267  & 604  \\
        \midrule[0.3pt]
        Stakeholder Feature Perception  & User Sentiment Analysis  & 3326  & 4000  \\       
             & Financial Report Analysis  & 19681 & 2000  \\ 
        \midrule[0.3pt]
        Real-time Market Discernment  & Stock Price Prediction  & 5510  & 4049 \\
             & Portfolio Asset Allocation  & --    & --   \\
        \midrule[0.5pt]
        \textbf{All}       
             & --  & --    & \textbf{28860} \\ 
        \bottomrule[1.2pt]
    \end{tabular}
\end{table*}

\begin{table*}[ht]
\centering
\caption{Details of the models evaluated in this paper. The "Access" column shows whether we have full access to the model weights or we can access through API. The “Version Date” column shows the release date of the corresponding version of the model we evaluated.}
\label{tab:model_details}
\vspace{10pt}
\resizebox{0.9 \textwidth}{!}{
\begin{tabular}{llccccc}
    \toprule[2pt]
    \textbf{Category} & \textbf{Model} & \textbf{Creator} & \textbf{Parameter} & \textbf{Access} & \textbf{Version Date} & \textbf{Domain}\\
    \midrule[1pt]
    Proprietary & GPT-5 & OpenAI & Undisclosed & API & 2025.11 & General\\
    & Gemini-3 & Google & Undisclosed & API & 2025.11 & General\\
    & Kimi-k2 & MoonshotAI & Undisclosed & API & 2025.11 & General\\
    & Claude-Sonnet-4 & Anthropic & Undisclosed & API & 2025.9 & General\\
    & Doubao-Seed-1.6 & ByteDance & Undisclosed & API & 2025.6 & General\\
    & Grok-4 & X AI & Undisclosed & API & 2025.7 & General\\
    & Qwen3-Max & Alibaba Cloud & Undisclosed & API & 2025.4 & General\\
    \midrule[1pt]    
    Open-Source & Qwen2.5-7B-Instruct & Alibaba Cloud & 7B & Weights & 2024.9 & General\\
    & Qwen2.5-72B-Instruct & Alibaba Cloud & 72B & Weights & 2024.9 & General\\
    & Qwen3-32B & Alibaba Cloud & 32B & Weights & 2025.4 & General\\
    & Qwen3-235B-A22B-Thinking-2507 & Alibaba Cloud & 235B & Weights & 2025.4 & General\\
    & InternLM2.5-7B & Shanghai AI Laboratory & 7B & Weights & 2025.3 & General\\
    & InternLM2.5-20B & Shanghai AI Laboratory & 20B & Weights & 2025.3 & General\\
    & GLM-Z1-9B & ZhipuAI & 9B & Weights & 2025.4 & General\\
    & GLM-Z1-32B & ZhipuAI & 32B & Weights & 2025.4 & General\\
    & DeepSeek-R1-Distill-Qwen-7B & DeepSeek AI & 7B & Weights & 2025.2 & General\\
    & DeepSeek-R1-Distill-Qwen-32B & DeepSeek AI & 32B & Weights & 2025.2 & General\\
    & DeepSeek-R1 & DeepSeek AI & 671B & Weights & 2025.12 & General\\
    & Fin-R1 & Shanghai University of Finance and Economics & 7B & Weights & 2025.3 & Financial\\
    & FinX1 & Duxiaoman-DI & 70B & Weights & 2024.12 & Financial\\
    & Dianjin-R1 & Alibaba Cloud & 32B & Weights & 2025.4 & Financial\\
    & Fino1 & The Fin AI  & 14B & Weights & 2025.2 & Financial\\
    \bottomrule[2pt]
\end{tabular}}

\end{table*}

\end{document}